\documentclass[a4paper,conference]{IEEEtran}
\usepackage{cite}

\ifCLASSINFOpdf
  \usepackage[pdftex]{graphicx}
\else
\fi
\usepackage{pgfplots}
\usepackage{booktabs}
\usepackage[absolute,overlay]{textpos}
\usepackage{multirow}
\usepackage{tikz}
\usetikzlibrary{arrows}

\usepackage{amsmath,amssymb} 
\usepackage{algorithm}
\usepackage[noend]{algpseudocode}
\makeatletter
\def\BState{\State\hskip-\ALG@thistlm}
\makeatother
\begin{document}
%
\title{Transferable Model for Shape Optimization subject to Physical Constraints}

\author{\IEEEauthorblockN{Lukas Harsch}
\IEEEauthorblockA{Institute of Fluid Mechanics\\ and Hydraulic Machinery\\
University of Stuttgart\\
Stuttgart, Germany 70569\\
lukas.harsch@ihs.uni-stuttgart.de}
\and
\IEEEauthorblockN{Johannes Burgbacher}
\IEEEauthorblockA{Institute of Fluid Mechanics\\ and Hydraulic Machinery\\
University of Stuttgart\\
Stuttgart, Germany 70569\\
johannesburgbacher@gmail.com}
\and
\IEEEauthorblockN{Stefan Riedelbauch}
\IEEEauthorblockA{Institute of Fluid Mechanics\\ and Hydraulic Machinery\\
University of Stuttgart\\
Stuttgart, Germany 70569\\
stefan.riedelbauch@ihs.uni-stuttgart.de}}


%


\maketitle

\begin{abstract}
The interaction of neural networks with physical equations offers a wide range of applications. We provide a method which enables a neural network to transform objects subject to given physical constraints. Therefore an U-Net architecture is used to learn the underlying physical behaviour of fluid flows. The network is used to infer the solution of flow simulations, which will be shown for a wide range of generic channel flow simulations. Physical meaningful quantities can be computed on the obtained solution, e.g. the total pressure difference or the forces on the objects. A Spatial Transformer Network with thin-plate-splines is used for the interaction between the physical constraints and the geometric representation of the objects. Thus, a transformation from an initial to a target geometry is performed such that the object is fulfilling the given constraints. This method is fully differentiable i.e., gradient informations can be used for the transformation. This can be seen as an inverse design process. The advantage of this method over many other proposed methods is, that the physical constraints are based on the inferred flow field solution. Thus, we have a transferable model which can be applied to varying problem setups and is not limited to a given set of geometry parameters or physical quantities.
\end{abstract}


%
\IEEEpeerreviewmaketitle

\section{Introduction}
With the great success of deep learning in various fields, there is an increasing interest to apply deep learning to physical problems. This implies learning physical equations, which can be used as a model to control robots \cite{planning_trajectories} or to solve numerical problems \cite{Raissi2018DeepHP, SIRIGNANO20181339}. In this work we propose a method to first predict the flow field for a given physical setup. Based on this, the shape of an object is estimated from physical quantities calculated on the predicted flow field. This framework for instance can be used to design airfoils for hydraulic machines. For a given set of fluid dynamical properties, e.g. forces on the airfoil surfaces, the system can estimate a shape of the airfoil to match these properties.\\
\indent
Especially the simulation of fluid flows using numerical solvers is a challenging task, which requires a strong understanding of the problem domain and complex mathematical models. In general, such physical systems can be characterized by a system of coupled non-linear partial differential equations (PDEs). In recent years several deep learning approaches \cite{DBLP:journals/corr/abs-1810-08217, 10.5555/3305890.3306035, Xie2018tempoGANAT} have shown promising results in the field of inference of fluid flow simulations in terms of computational time and prediction accuracy.\\
\indent
These approaches \cite{Lagaris_1998, Raissi2018DeepHP, Aarts_2001, SIRIGNANO20181339} are either designed to solve a specific PDE for a defined problem subject to boundary conditions to infer the temporal evolution of the system. Or they are designed as data driven models \cite{10.5555/3305890.3306035, Xie2018tempoGANAT, schenck2018spnets, Ummenhofer2020Lagrangian}, which aim to learn the underlying physical behaviour of fluid dynamics from data. These models can be used to infer the flow field (e.g. estimating the velocity field or fluid motion) for another problem setup. The data-driven models in general differ with respect to their representation of the fluid. This includes the representation of the flow field on a Cartesian grid so that image processing algorithms can be applied \cite{DBLP:journals/corr/abs-1810-08217, 10.1145/2939672.2939738}. In common numerical solvers the flow field is represented as a mesh of arbitrary polygons, which can be processed with graph neural networks \cite{wu2019comprehensive}. Further, the fluid can be treated as particles, which can be seen as a point cloud \cite{schenck2018spnets, Ummenhofer2020Lagrangian}.
\begin{figure}[t]
	\center
	\def\svgwidth{250pt}    
	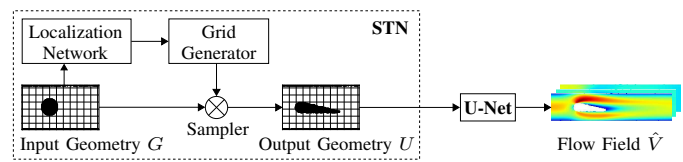  
	\caption{Combined model with a pretrained U-Net to predict the flow field for a given input geometry and a Spatial Transformer Network (STN) in front to optimize the input geometry subject to physical constraints calculated on the predicted flow field.}
	\label{combinedModel}
\end{figure}\\
\indent
In this work we focus on a data-driven model on a Cartesian grid. To train the model several hundred CFD-simulations of a channel flow with objects of different shapes and order inside the channel are used as true data. In a preprocessing step they get interpolated onto a regular grid, so that a convolutional neural network (CNN) can be used to process the data. The main goal of this work is to connect physical equations that depend on the flow field with a neural network and show how the input geometry can be manipulated with respect to these equations.\\
\indent
In a first step a well established system \cite{DBLP:journals/corr/abs-1810-08217} based on a U-Net \cite{DBLP:journals/corr/RonnebergerFB15} architecture is used to learn the fluid dynamics and to infer the flow field for a given new geometry setup. On the basis of these solutions we are able to calculate meaningful equations which describe the system, e.g. the total pressure difference or the forces on the objects. These equations are further used as physical constraints for the shape optimization.\\
\indent
In a second step a \textit{Spatial Transformer Network}~(STN)~\cite{DBLP:journals/corr/JaderbergSZK15} is used to manipulate the input object to control the equations of interest. Since this method is fully differentiable the gradients of the equations calculated on the out coming flow field can be propagated back to the input geometry. With this gradient information the STN is able to change the shape of the object according to the given equations.\\
\indent
For the optimization a two-step training process is used. First, the U-Net gets trained in isolation to learn the inference of the flow field. Second, the U-Net and the STN act as a combined model, where the parameters of the U-Net are frozen. Thus, only the transformation of the input shape will be optimized, while the performance of the flow field prediction will be kept constant. The combined model is depicted in Figure \ref{combinedModel}.\\
\indent
This method can be seen as a kind of inverse design process which in general presents major challenges in terms of numerical algorithms, computational time and convergence. From the forward pass, a flow field is obtained where the specified constraints can be calculated. Afterwards in the backward step the input shape gets transformed to fulfil these constraints.\\
\indent
Besides using a STN to transform the shape it is also possible the use a GAN \cite{NIPS2014_5423} architecture to generate the optimal shape from a noise vector. Nevertheless this method can lead to scattered geometries over the whole channel. With the STN it is possible to prevent this by using a properly initialized geometry that retains the transformed geometry as a solid object.\\
\indent
%
In contrast to the presented approach, other inverse design methods \cite{SUN2015415, Nanophotonic_Structures} are using a direct mapping from a fixed set of geometry parameters to given physical quantities. These methods require retraining and new datasets in case of changing geometry parameters or physical quantities. These methods may outperform the presented approach for a fixed setup but require retraining and new datasets in case of changing geometry parameters or physical quantities. So these models are not flexible. Our proposed method is more general, since it can reuse the estimation of the flow field in case of a changing geometry setup and is not limited by predefined set of geometry parameters. Based on this solution varying physical quantities can be calculated. Thus, we have a transferable model which can be applied to a wide range of problem configurations.\\
\\
\indent
The contributions of our work are summarized below:
\begin{itemize}
 \item A fully differentiable flow prediction framework based on a U-Net architecture in combination with a STN is proposed to optimize the shape of the input geometry with respect to physical constraints calculated on the predicted flow field.
 \item The framework is successfully applied to generic objects such as rectangles, triangles and circles, and more realistic objects such as airfoils for different physical equations implying the total pressure difference, object forces, geometric restrictions and the full flow field.
 \item Our method relies on predicting the flow field in a first step and hence can be used for varying physical quantities and geometry parameters. Thus, a transferable model is developed for a wide range of setups. 
 \item Our dataset is available at \textbf{https://github.com/Flow-Field-Prediction/2D-Channel-Flow} for experiments and further studies.
\end{itemize}
%
%
%
%
%
\section{Related Work}
There are a few approaches which can be used to predict the flow field for a given setup. A major group is approximating the unknown solution of PDEs to satisfy the boundary and initial conditions. Scattered observation with spatial and temporal coordinates~($x$, $t$) are used as input to predict the corresponding output values~$u=f(x,t)$. To learn the underlying PDE, the solution of the PDE~$J(f(x,t))$ is used as loss function, thus~$J(f)$ measures how well~$f(x,t)$ satisfies the PDE. The concepts mainly differ on how to approximate the derivatives of the PDEs, e.g.~$\frac{\partial u}{\partial x}$, $\frac{\partial u}{\partial t}$. The ability of a fully differentiable network is used in \cite{Lagaris_1998, Raissi2018DeepHP, raissi2018hidden} to propagate the predicted output~$u$ back to the inputs~$(x,t)$ to get the derivatives. In \cite{Aarts_2001} separate linear layers are used to approximate the particular derivatives and connect them to model the solution of the PDE. In contrast, the first order derivatives can be calculated analytically and a Monte-Carlo method is used to approximate the second order derivatives~\cite{SIRIGNANO20181339}. These methods can be used to infer the solution for a further time horizon or to learn a general solution for a wide range of boundary, initial and physical conditions.\\
\indent
Another type of approaches does not use the underlying PDEs, instead the fluid dynamics are learned from a large amount of data with the use of certain distance measures between the predicted and the true flow field. Thus, a model should be trained to learn a generalized solution of the fluid dynamics to infer the solution of the flow field for new setups. In \cite{10.1145/2939672.2939738, DBLP:journals/corr/abs-1810-08217} the data is stored on a regular two-dimensional grid to be processed with common convolutional neural networks~(CNN) by optimizing the mean square error (MSE). These very basic concepts are used in this work for the flow field prediction, since we focus on the shape optimization. The prediction of the flow field can be replaced by more complex methods in a further work. An extension onto the three-dimensional case is presented in \cite{10.5555/3305890.3306035, Kim2018DeepFA, Latent_Space_Physics, Xie2018tempoGANAT}. These methods also provide the ability of learning the temporal evolution of fluid flows.\\
\indent
Besides the representation of data as a regular grid, the fluid can also be treated as particles and stored as a point cloud. In \cite{schenck2018spnets, Ummenhofer2020Lagrangian} new types of convolution layers are introduced to process point clouds to predict fluid motion. These models are further able to adapt fluid parameters to moving objects and boundaries.\\
\indent
In the past several methods for inverse design of geometries were presented. In \cite{doi:10.2514/1.J057894} the graph of the pressure distribution~$p(x)$ on the surface of an airfoil over its \mbox{$x$-coordinates} is used as input for a CNN to predict the corresponding \mbox{$y$-coordinates} of the target airfoil shape. Another approach~\cite{SUN2015415} uses a neural network (NN) to link a large amount of aero-dynamic quantities with geometry data. They use different parametrizations to represent the geometry information. 
In \cite{Nanophotonic_Structures} structural parameters~$D$ are transformed to electromagnetic responses~$\hat{R}$. First a model is trained to map from~$D$ to~$\hat{R}$, then freezing the weights. Afterwards another network is stacked in front of it to learn the mapping from~$R$ to~$D$ to~$\hat{R}$, while minimizing the~$MSE(R,\hat{R})$. Thus, using~$R \rightarrow D$ in isolation the structural parameters~$D$ for a specified electromagnetic response~$R$ can be predicted. We also use a pretrained model for the mapping from geometry information to physical response, but extend it further, as we iteratively transform the geometry to obtain a direct response in terms of the physical quantities defined over the entire fluid field.
It has to be mentioned that these methods are using a fixed predefined set of input features and corresponding geometric representations. Thus, a new dataset and retraining is required in the case of changing features, so they are not directly comparable to our approach. In contrast our approach has a transferable model and can be used for changing problem configurations. 
\section{Methodology}
In this section we present an overview of the proposed method for learning the fluid dynamics and predicting the flow field for new setups. Afterwards the Spatial Transformer Network (STN) is presented to change the shape of the input objects. These concepts are then coupled to propagate gradients of the loss function calculated on the predicted flow field back to the input object. Thus, it is possible to change its shape with respect to physical quantities defined by the loss function. In the last step different physical quantities and their associated equations are highlighted.
\subsection{Dataset}
\begin{figure}[t]
\center
\begin{tabular}{cc}
  \includegraphics[width=40mm]{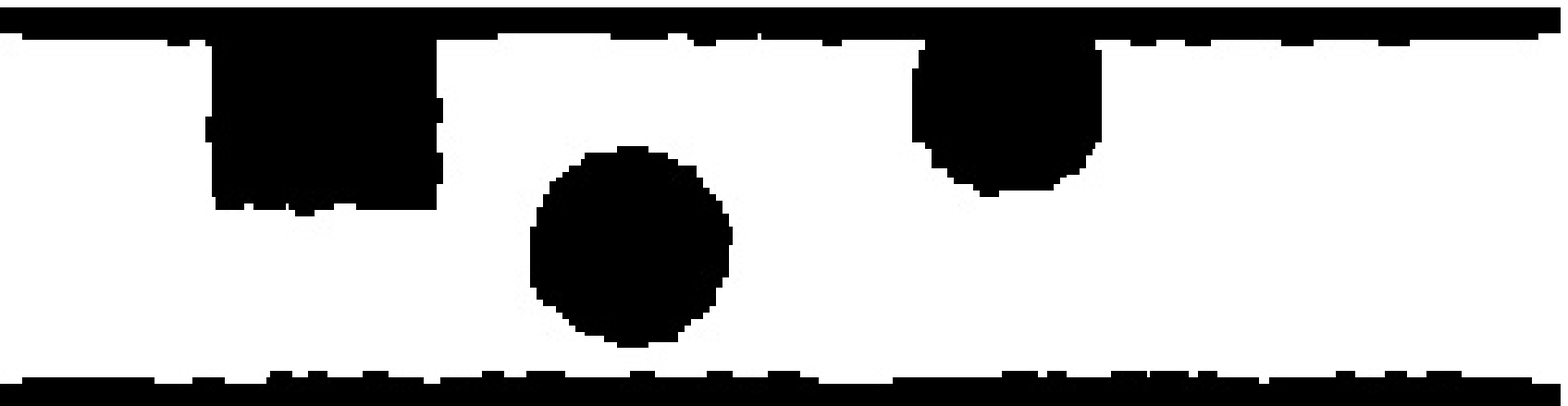} & \includegraphics[width=40mm]{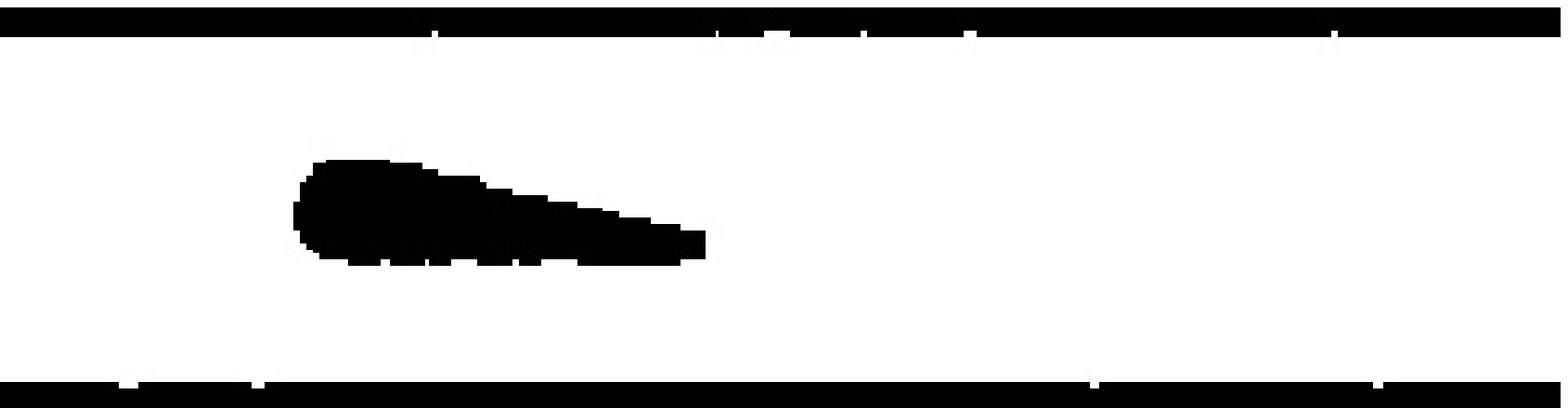}\\
  \includegraphics[width=40mm]{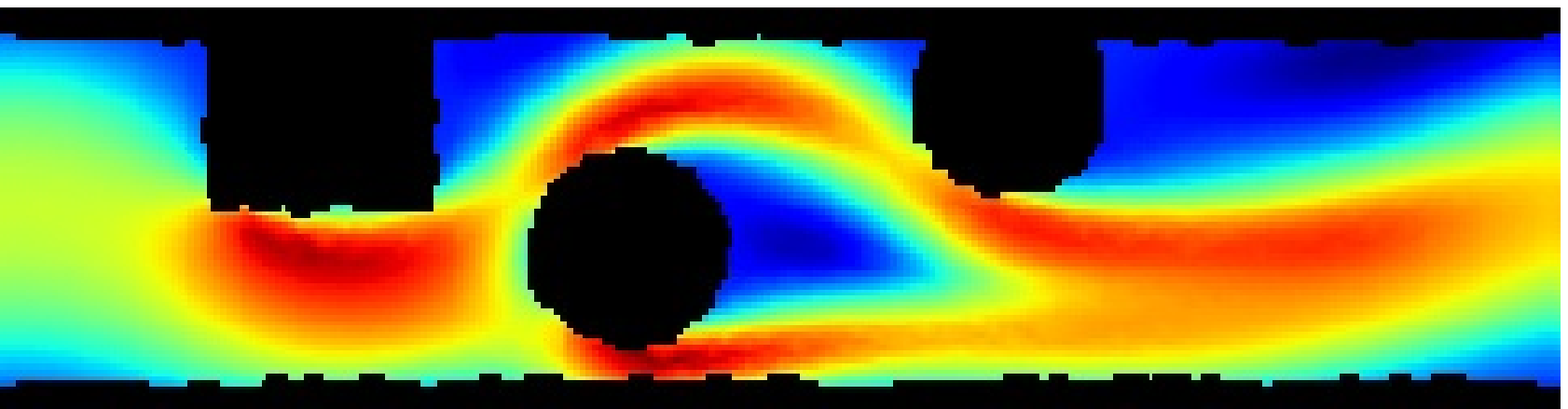} &  \includegraphics[width=40mm]{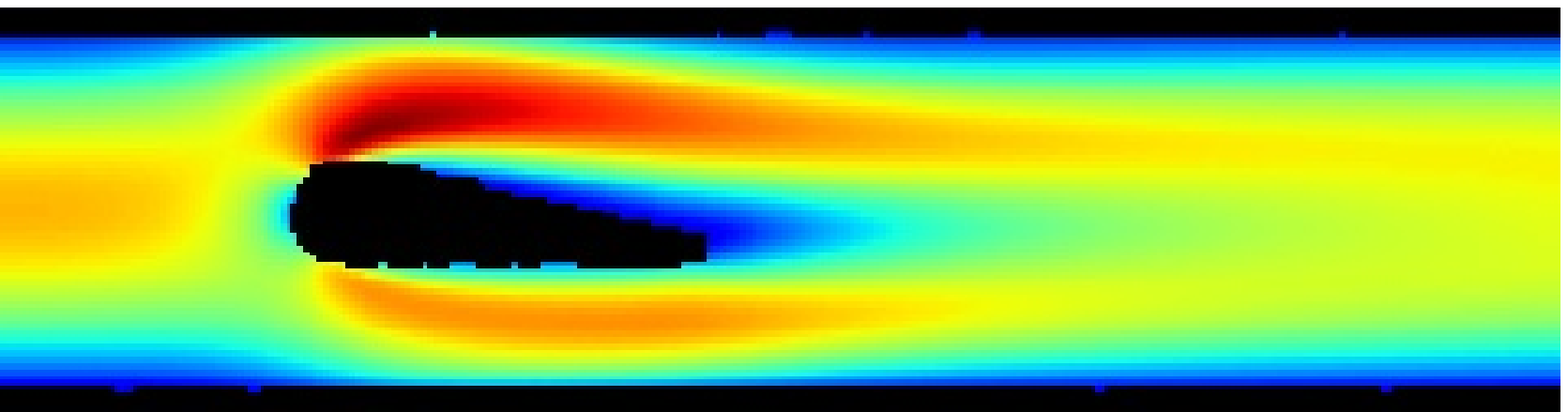}\\
  \includegraphics[width=40mm]{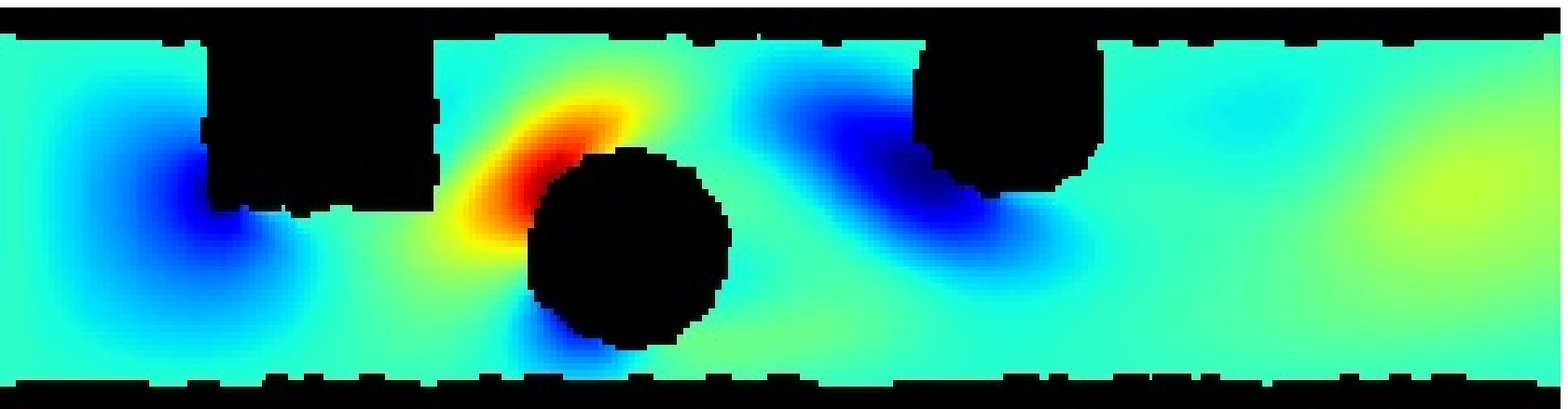} & \includegraphics[width=40mm]{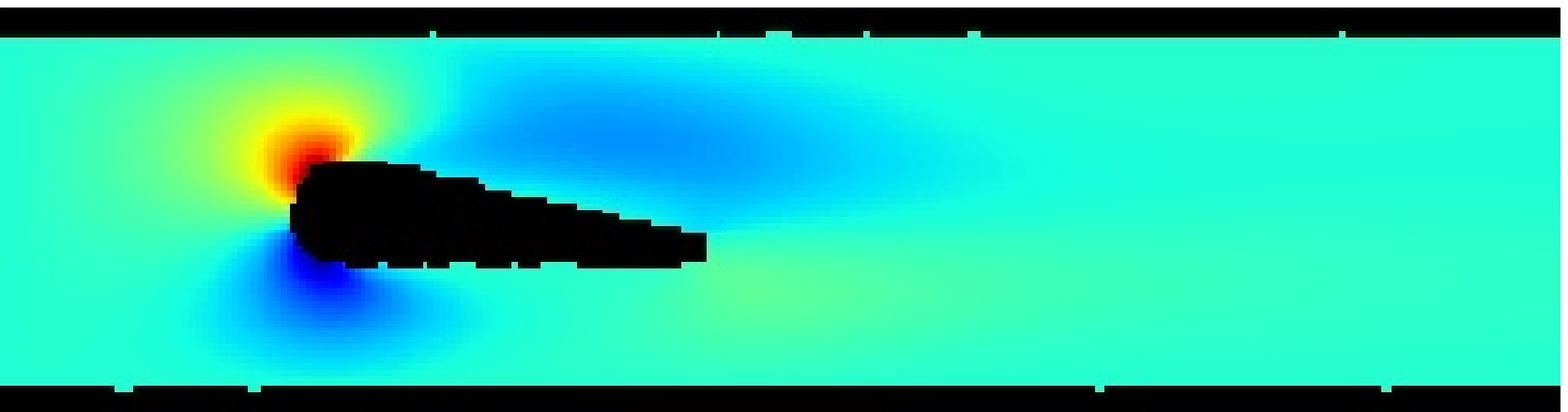}\\
  \includegraphics[width=40mm]{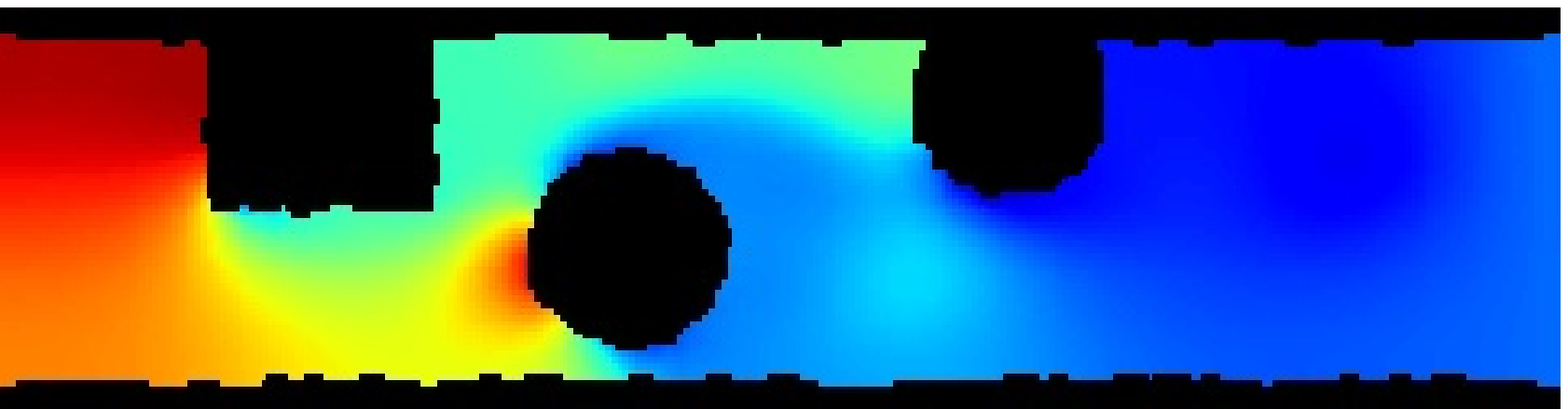} &  \includegraphics[width=40mm]{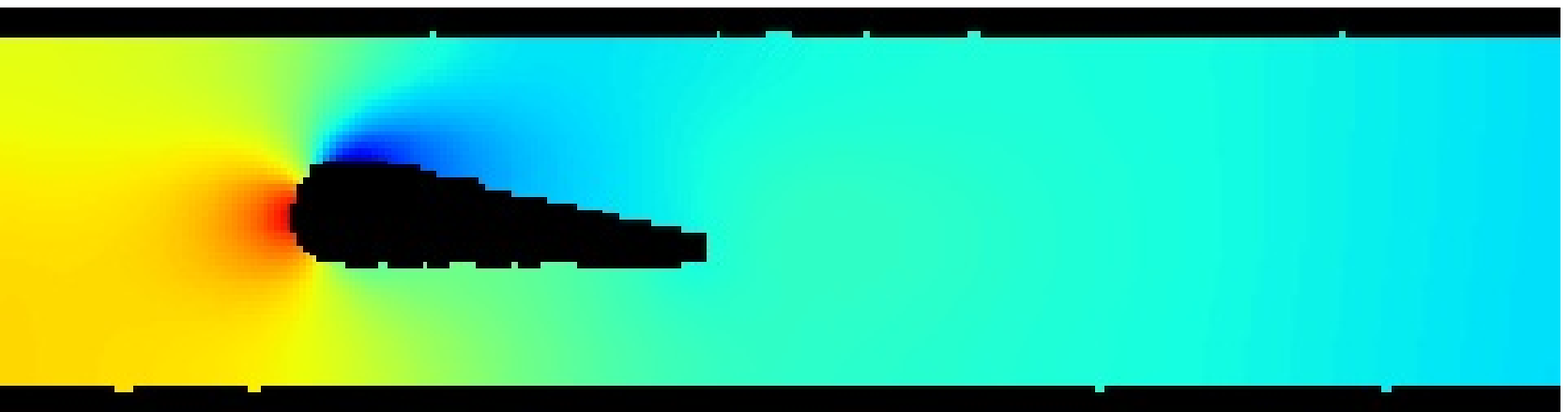}\\
\end{tabular}
\caption{Samples from the dataset interpolated to images. \textit{Top}:~Geometry. \textit{Bottom}: Flow field with three channels $u$,~$v$,~$p$.}
\label{dataset}
  \begin{textblock}{0.5}(1.13,1.68)
    $u$
  \end{textblock}
  \begin{textblock}{0.5}(1.13,2.31)
    $v$
  \end{textblock}
  \begin{textblock}{0.5}(1.13,2.935) 
    $p$
  \end{textblock}
  \begin{textblock}{0.5}(4.5,1.68)
    $u$
  \end{textblock}
  \begin{textblock}{0.5}(4.5,2.31)
    $v$
  \end{textblock}
  \begin{textblock}{0.5}(4.5,2.935) 
  $p$
  \end{textblock}
\end{figure}
The simulation of fluid flow is a challenging task since it requires a complex mathematical model to capture changing motions and boundary conditions in a complex domain. Such systems are in general modelled as a system of coupled non-linear PDEs. An established physical formulation to describe fluids are the \textit{Navier-Stokes Equations}~(NSE)
\begin{equation}
 \frac{\partial}{\partial t}(\rho \textbf{u}) + \rho \textbf{u} \cdot \nabla \textbf{u} = - \nabla p + \mu \nabla^2 \textbf{u} + \rho \textbf{g},\;\;
 \nabla \cdot \textbf{u} = 0,
\end{equation}
with fluid velocity~$\textbf{u}$, pressure~$p$, external forces~$\textbf{g}$, density~$\rho$ and dynamic viscosity~$\mu$. To solve this system of equations numerical methods such as finite-differences, finite-volumes or finite-elements are used. These methods require a discretization of the domain. Common techniques are grid-like structures to model the interaction inside the fluid.\\ 
\indent
To learn the fluid dynamics with a neural network, a large amount of fluid simulations is required as training data. In this work the simulations are done with FEniCS \cite{AlnaesBlechta2015a}, which uses a finite-element method to solve the PDEs. FEniCS has an automatic process to mesh the domain with unstructured triangular elements. Every node in the mesh holds values for the velocity in $x$- and $y$-direction~($u,v$) and the pressure~$p$, all together representing the flow field. To use this as input for a convolutional neural network, the point set of nodes of the unstructured mesh are linearly interpolated to a Cartesian grid~$V \in \mathbb{R}^{H\mathrm{x}W\mathrm{x}C}$. This represents the three values~$u$,~$v$~and~$p$ of the field with height~$H=64$, width~$W=256$ and number of channels~$C=3$. Each element of the Cartesian grid can be interpreted as a pixel, thus common convolutional layers can be used to process these data. The size of $V$ is chosen such that the original aspect ratio of the channel is kept and enough details of the fluid flow are captured. Two examples for the interpolated flow fields are shown in Figure \ref{dataset} with the three channels $u$, $v$ and $p$, respectively. The parts in the flow field where the geometry is placed are highlighted in black. In the data matrix $V$ these parts are filled with zeros. The flow field on the right side is showing a typical behaviour for an airfoil with high velocity components $u$ in $x$-direction on the top side and lower velocities below which results in a low pressure area above the airfoil.\\
\indent
Before a fluid simulation can be done, the domain has to be modelled i.e., the walls and solid objects inside the domain have to be defined. In this work a 2D~channel flow is used as simulation domain, which has a wall at the top and the bottom and some immersed objects inside the channel. This geometry information stored on the unstructured mesh is also interpolated on a Cartesian grid~$G\in \mathbb{R}^{H\mathrm{x}W\mathrm{x}1}$ as a geometry map with~$C=1$, where a~0 indicates fluid and a~1 represents solid objects. The linear interpolation is leading to blurry border regions between the objects and the fluid, which are rounded to 0 and 1. All simulations in the dataset are stationary 2D~channel flows with a varying number of objects~$n\le3$, randomly placed inside the channel. Objects can be simple geometric forms such as squares, rectangles and triangles as well as more complex shapes such as airfoils.\\
\indent
The dataset~$D=\{(G^i,V^i)|i=1,\ldots,N\}$ consists of pairs of geometry~$G$ and flow field~$V$ with a total amount of~$N=1350$ samples. Thus, the geometry~$G$ is used as input for the network, while the flow field~$V$ is the ground truth for training the model in a supervised fashion. The data is split to have 1050~samples for training the model and 300~samples for evaluation. 
\subsection{Flow Field Prediction}
\begin{table}[t]
\begin{center}
\caption{The different stages of the U-Net architecture. The notation ''3x3 Conv(stride=1,pad=1)x$C$ - leakyReLU(0.1)'' denotes a convolutional kernel with filter size~3x3 and $C$~output channels, followed by a leakyReLU activation with slope~${s=0.1}$.}
\label{UNet_architectur}
 \begin{tabular}{l l} 
 \hline
 \multicolumn{1}{l}{Layer Name} & \multicolumn{1}{c}{Settings}\\ 
 \hline\hline
 Input & Geometry Map ${H\mathrm{x}W\mathrm{x}1}$\\ 
 \hline
 Comp. Stage & \begin{tabular}{@{}l@{}}3x3 Conv(stride=1,pad=1)x$C$ - leakyReLU(0.1) \\
					     3x3 Conv(stride=1,pad=1)x$C$ - leakyReLU(0.1) \\
					     2x2 Pooling(stride=2,pad=0)\end{tabular}\\
 \hline
 Bottom Stage & \begin{tabular}{@{}l@{}}3x3 Conv(stride=1,pad=1)x$C$ - leakyReLU(0.1) \\
					     3x3 Conv(stride=1,pad=1)x$C$ - leakyReLU(0.1)\end{tabular}\\
 \hline
 Exp. Stage & \begin{tabular}{@{}l@{}}2x2 TranspConv(stride=2,pad=2)x$C$ - leakyReLU(0.1) \\ 
				      Concatenate with corresponding Comp. Stage\\
				      3x3 Conv(stride=1,pad=1)x$C$ - leakyReLU(0.1)\\
				      3x3 Conv(stride=1,pad=1)x$C$ - leakyReLU(0.1)\end{tabular}\\
 \hline
 Output & Flow Field ${H\mathrm{x}W\mathrm{x}C}$\\
 \hline
\end{tabular}
\end{center}
\end{table}
\begin{table}[t]
	\begin{center}
	\caption{Architecture of the Localization Network.}
	\vspace{-1\baselineskip}
	\label{LocNet_architecture}
		\begin{tabular}{l l} 
			\hline
			\multicolumn{1}{l}{Layer Name} & \multicolumn{1}{c}{Settings}\\
			\hline\hline
			Input & Geometry Map ${H\mathrm{x}W\mathrm{x}1}$\\ 
			\hline
			Conv1 & 3x3 Conv(stride=2,pad=1)x64 - leakyReLU(0.2)\\
			\hline
			Conv2, Conv3, Conv4 & 3x3 Conv(stride=2,pad=1)x128 - leakyReLU(0.2)\\
			\hline
			Conv5 & 3x3 Conv(stride=2,pad=1)x64 - leakyReLU(0.2)\\
			\hline
			FC1 & Linear (2054 units) - Tanh\\
			\hline
		\end{tabular}
	\end{center}
\end{table}
To highlight the strong capability of the model in learning the behaviour of fluid flow, no physical prior on the input data such as boundary condition for the velocities and pressure or conservation laws is used. Thus, we simply try to find a mapping~$f(G)=\hat{V}$ from the input geometry~$G$ to the corresponding fluid field~$\hat{V}$. Previous work \cite{DBLP:journals/corr/abs-1810-08217} has already shown strong results in using a U-Net architecture \cite{DBLP:journals/corr/RonnebergerFB15} to learn the dynamics of the fluid from data. We also rely on the \mbox{U-Net} for predicting the flow field, but using a dataset with a large variety of fluid dynamics i.e., various flows, backflows, partial blockage and small gaps. Thus, the network has to capture a wide range of phenomena which makes it challenging to train a suitable model. It can be shown that this method offers promising results regarding the overall error in~($u$,$v$,$p$) and with respect to further analysis of the flow field. This is essential since the main focus of this work is to couple the flow field prediction with a shape optimization subject to physical constraints.\\
\indent
The U-Net is a fully convolutional network described in Table~\ref{UNet_architectur}. Five compression stages are used to reduce the size of the input data, where each has two convolution layers with constant kernel size of~3x3 and doubling channel size~$C=[32,\ldots,512]$ followed by a max-pooling layer and one bottom stage without a pooling layer. Afterwards five expansion stages are applied to get back to the original dimension of the input data. A transposed convolution with a kernel size of~2x2 is used to increase the size of the data with a striding of~2x2. To this the output of the corresponding compression stage is concatenated, followed by two standard convolution layers. The channel size~$C=[512,\ldots,3]$ will be halved per stage, except in the last stage. The activation function leakyReLU with slope~$s=0.1$ is applied, except in the last layer, where the hyperbolic tangent should transform the output to a proper scale. The network is trained with the Adam-optimizer~\cite{kingma2014adam} with a learning rate of~$10^{-4}$ and the sum of squared errors~(SSE) as loss function
\begin{equation}
 \mathcal{L}_{SSE}=\sum_{i=0}^{N}(\hat{V}_i-V_i)^2\;,
\end{equation}
between the predicted flow field $\hat{V}$ and the true flow field~$V$ from the simulations. The SSE is used as loss function to have a strong restriction on every pixel, instead of getting a low average error.
\subsection{Geometry Transformation}
With the presented U-Net architecture it is possible to predict the flow field for a new geometry setup. Since the goal is an interaction of the flow field with a shape optimization subject to physical constraints, it is required to deform the input geometry with respect to the deployed constraints. Therefore a Spatial Transformer Network (STN) \cite{DBLP:journals/corr/JaderbergSZK15} is used.\\
\indent
The STN is a three-steps method. In the first step a localization network~$\theta = f_\mathrm{Loc}(G)$ takes the input geometry map~$G\in \mathbb{R}^{H\mathrm{x}W\mathrm{x}1}$ to compute the parameters~$\theta$ for the transformation~$\tau _\theta$. Depending on the parametrization of the transformation, e.g. an affine transformation, the size of~$\theta$ can vary.\\
\indent
In the next step a parametrized sampling grid is required. The input geometry map is a regular grid~$R={R_i}$ which is a set of points with source coordinates~$R_i=(x^s_i,y^s_i)$. Each element in~$R_i$ defines a pixel in the input geometry map. Then the transformation~$\tau_\theta$ can be applied on~$R$
\begin{equation}
 \begin{bmatrix} x^t_i \\ y^t_i \end{bmatrix}
 = \tau_\theta (R_i) = \mathrm{A}_\theta
 \begin{bmatrix} x^s_i \\ y^s_i \\ 1 \end{bmatrix}\;,
\end{equation}
which puts out a set of points with target coordinates~$(x^t_i,y^t_i)$. The matrix~$\mathrm{A}_{\theta}$ denotes the affine transformation matrix. The height and width of the source coordinates are normalised, such that~$x^s_i, y^s_i \in [-1,1]$. The same interval holds for the target coordinates in the output domain.\\
\indent
In the third step a set of sampling points~$\tau_\theta(R)$ together with the input geometry map~$G$ can then be used to generate a transformed output geometry map~$U\in \mathbb{R}^{H'\mathrm{x}W'\mathrm{x}1}$. This is done with a bilinear sampling kernel of the form:
\begin{equation}
 U^c_i = \sum_{n,m}^{H,W} G^c_{nm} \mathrm{max}(0,1-|x^s_i-m|)\mathrm{max}(0,1-|y^s_i-n|)\;,
\end{equation}
for~$\forall i \in [1,\ldots,H'W']$ and~$\forall c\in [1,\ldots,C]$ to get the output value of every element~$U_i^c$ at the location~$(x^t_i,y^t_i)$ for every channel~$C$.\\
\indent
To deform the input geometry with respect to physical constraints, complex transformations may be necessary. Therefore, the usage of an affine transformation is not sufficient. In this work thin-plate-splines (TPS) \cite{24792, 10.1007/3-540-47977-5_2} are used as a more complex parametrization of the transformation.
\begin{algorithm}[t]
 \begin{algorithmic}[1]
   \State \textbf{Input:}
   \State Train dataset $D=\{(G^i,V^i)|i=1,\ldots,N\}$
   \State New initial geometry map $G_{init}\in \mathbb{R}^{H\mathrm{x}W\mathrm{x}1}$   
   \Procedure{Train U-Net}{$G,V$}
   \For {$n$ epochs} {}
   \State Predict Flow Field $\hat{V} = f_{UNet}(G)$ 
   \State Update $\theta_{\mathrm{UNet}}$ $\leftarrow$ Loss $\mathcal{L}_{SSE}(\hat{V},V)$
   \EndFor
   \State \textbf{return} U-Net parameters $\theta_{\mathrm{UNet}}$
   \EndProcedure

   \Procedure{Initialize Loc-Net}{$G$}
   \For {$n$ epochs} {}
   \State Parameters for Transformation $\theta = f_{Loc}(G)$
   \State Grid Generator $(x^t,y^t)=\mathrm{TPS}(x^s,y^s,\theta)$
   \State Transform Geometry $U=\mathrm{Sampler}(x_t,y_t,G)$ 
   \State Update $\theta_{\mathrm{Loc}}$ $\leftarrow$ Loss $\mathcal{L}_{SSE}(U,G)$
   \EndFor
   \State \textbf{return} Loc-Net parameters $\theta_{\mathrm{Loc}}$
   \EndProcedure
  
   \Procedure{Transformation}{$G_{init}$, $\theta_{\mathrm{UNet}}$, $\theta_{\mathrm{Loc}}$}
   \For {$n$ epochs} {}
   \State Transformation parameters $\theta = f_{Loc}(G_{init})$
   \State Grid Generator $(x^t,y^t)=\mathrm{TPS}(x^s,y^s,\theta)$
   \State Transform Geometry $U=\mathrm{Sampler}(x_t,y_t,G_{init})$ 
   \State Predict Flow Field $\hat{V} = f_{UNet}(U)$ 
   \State Update $\theta_{\mathrm{Loc}}$ $\leftarrow$ Physical Constraints $\mathcal{L}_{total}(\hat{V})$
   \EndFor
   \State \textbf{return} Transformed Geometry $U$
   \EndProcedure
 \end{algorithmic}
 \caption{Training procedure for geometry transformation}
 \label{pseudoCode}
\end{algorithm}
\\
\indent
The TPS interpolation~$\mathrm{TPS}(x,y)$ is defined over a set of $p$~control points with coordinates~$(x^c_i, y^c_i)$, $i=1,\ldots,p$ which 
are chosen among the points of the regular grid~$R={R_i}$ with source coordinates~$(x^s_i, y^s_i)$. The function values~$\mathrm{TPS}(x^s_i, y^s_i)$ then corresponds to target coordinates~$(x^t_i,y^t_i)$. The TPS interpolation has the form
\begin{equation}
\begin{split}
\mathrm{TPS}(x^s,y^s)= &\; \theta_{p+1} + \theta_{p+2}\;x^s + \theta_{p+3}\;y^s \\
& +\sum_{j=1}^p \theta_j T(||(x^c_j, y^c_j) - (x^s,y^s)||)\;,
\end{split}
\end{equation}
where~$T(r)=r^2 \mathrm{log} r$. As for the affine transformation the TPS interpolation has a set of parameters~$\theta$ to control the transformation which are again the outputs of the localization network $f_\mathrm{Loc}$. Afterwards a bilinear sampling kernel as in the original STN paper is applied to get the transformed image.\\
\indent
For the localization network $f_\mathrm{Loc}$ a standard CNN is used, with one regression layer at the end. A detailed description of the network architecture can be found in Table~\ref{LocNet_architecture}. The output dimension of the localization network is depending on the number of parameters which are used for the TPS interpolation. In this work equally distributed control points with a step size of~4 in both spatial dimensions are used. Thus~$p=16\mathrm{x}64$~plus~3 parameters for every spatial dimension are required. 
\subsection{Combined Model}
The final model is a combination of the U-Net with the STN in front of it, whereas the U-Net predicts the flow field for a given input while the STN deforms the input repeatedly. The overall training procedure for the geometry transformation is summarized in Algorithm \ref{pseudoCode}. The training of this model is done in a two stage procedure. First the U-Net is trained in isolation to learn a mapping from input geometry $G$ to flow field $\hat{V}$. Afterwards the U-Net and the STN are used in combination. At this point the parameters~$\theta_{\mathrm{UNet}}$ of the U-Net are frozen and only the parameters~$\theta_{\mathrm{Loc}}$ of the STN are updated to perform the transformation of the input. Since this method is fully differentiable from the predicted flow field back to the input geometry, the STN learns how to transform the shape of the input to fulfil the physical constraints calculated on the output. With this procedure it is possible to get both a high precision of the predicted flow field and an interaction between the constraints and the geometrical shape.\\
\indent
It has to be mentioned that the STN requires a proper initialization. Therefore, the STN is pretrained to reproduce the input geometries from the training data set, which simply means to pass input through the STN without any transformation. This is important since otherwise the STN tends to diverge, because of strong transformation at the beginning of the training. Moreover it turns out that a learning rate of~$10^{-5}$ leads to the best optimization performance. With a small learning rate the shape is changing slightly so the STN can better react to changes in the flow field and hence in the loss function.  
\subsection{Physical Constraints}
\begin{figure}[t]
\center
  \def\svgwidth{200pt}    
  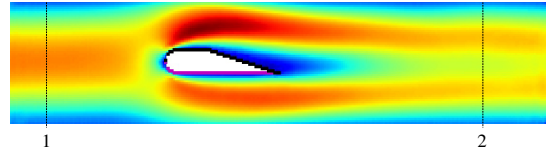  
\caption{Inlet and outlet boundary for calculating $\Delta p$ and the top and bottom surface of the airfoil used for the forces $F_i$ marked in black and magenta.}
\label{BoundariesForces}
\end{figure}
The overall goal of this work is the manipulation of input geometries with respect to physical equations, which are acting as constraints to the system. These equations are calculated from the values of the output flow field and are thus coupled to the input geometry via both networks. Two equations for constraining the optimization are presented. This includes the total pressure difference~$\Delta p$ between two boundaries
\begin{equation}
 \Delta p = \Big(\frac{\rho}{2} (|u_2+v_2|^2)+p_2\Big) - \Big(\frac{\rho}{2} (|u_1+v_1|^2)+p_1\Big)\;,
\end{equation}
with the velocities $u$, $v$, the pressure $p$ and the density of the fluid~$\rho$. The subscripts 1 and 2 denote the inlet and outlet boundaries between which $\Delta p$ is calculated, as shown in Figure~\ref{BoundariesForces}. For~$\Delta p$ the values $u$, $v$ and $p$ are averaged over the boundary area. Since all data are stored on a regular grid, the functions are evaluated on a discrete domain, i.e. per pixel. For example the velocity~$u_n$ used for $\Delta p$ is simply the $n$-th column in the data matrix of the channel which corresponds to the~$u$-values.
\\
\indent
Also the pressure forces~$F_i$ on the object surface are used as constraints for the shape optimization
\begin{equation}
 F_i = \int_{A_i} p_i\;dA_i\;,
\end{equation}
where~$A_i$, $i=1,\ldots,4$ represents the front, back, top and bottom surface and $p_i$ the pressure. Figure \ref{BoundariesForces} displays an example of the pixels used for the top and bottom surface, marked in black and magenta. In the discrete domain the integral of the force~$F_i$ can be written as a sum over all pressure values of the pixels belonging to the surface of the geometry. Since the surface~$A_i$ is constructed from a set of pixels with a constant edge length, this is simply multiplying the pressure~$p_i$ for every pixel by the edge length of the pixel.
\\
\indent
The total loss for the shape optimization is then given as
\begin{equation}
 \mathcal{L}_{\mathrm{total}} = (\Delta \hat{p} - \Delta p)^2 + \sum_{i=1}^4 (\hat{F}_i - F_i)^2\;.
\end{equation}
The true values~$\Delta p$ and~$F_i$ are calculated on the flow field for a reference geometry to which we want to optimize the shape. Indeed, many other functions can be calculated on the flow field and be used as loss function, depending on how the physical system should be restricted. We further show this by using the length~$L$, height~$H$ and centre of mass~$COM$ of the object as additional criterions for the optimization. 
\pgfplotsset{compat=1.5, 
	every tick label/.append style={font=\scriptsize},
	legend image code/.code={\draw[mark repeat=1,mark phase=1] plot coordinates {(0cm,0cm)(0.0cm,0cm)(0.1cm,0cm)};},
}
\begin{figure}[!t]
	\begin{tabular}{cc}
		\begin{tikzpicture}
		\begin{axis}[
		width=135pt, 
		height=100pt,
		xlabel={\scriptsize Channel Length},
		ylabel={\scriptsize Relative Error [\%]},
		xmin=0,
		xmax=255,
		ymin=0,
		ymax=8.5,
		legend pos=north west,
		grid=major,
		]
		\addplot[color=blue] table {relativErrors_u.txt};
		\addplot[color=red] table {relativErrors_v.txt};
		\addplot[color=green] table {relativErrors_p.txt};
		\addlegendentry{\scriptsize $u$}
		\addlegendentry{\scriptsize $v$}
		\addlegendentry{\scriptsize $p$}
		\end{axis}
		\end{tikzpicture}
		&
		\begin{tikzpicture}
		\begin{axis}[
		width=135pt, 
		height=100pt,
		xlabel={\scriptsize Channel Length},
		ylabel={\scriptsize Relative Error [\%]},
		xmin=0,
		xmax=255,
		ymin=0,
		ymax=8.5,
		legend pos=north west,
		grid=major,
		]
		\addplot[color=blue] table {relativErrors_mass.txt};
		\addplot[color=red] table {relativErrors_energy.txt};
		\addlegendentry{\scriptsize $\Delta Q$}
		\addlegendentry{\scriptsize $\Delta p$}
		\end{axis}
		\end{tikzpicture}
	\end{tabular}
	\caption{Distribution of the mean relative error along the channel length. \textit{Left}: Error in $u$, $v$ and $p$. \textit{Right}: Error in $\Delta Q$ and $\Delta p$.}
	\label{relativeErrors}
\end{figure}
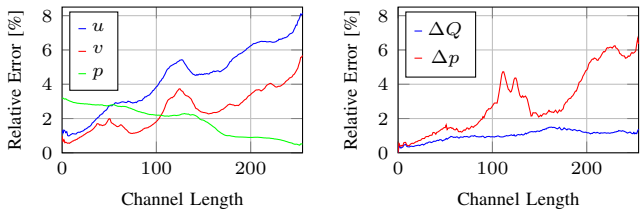
\begin{figure*}[t]
	\center
	%
	\begin{tabular}{ccccc}
		\includegraphics[width=20mm]{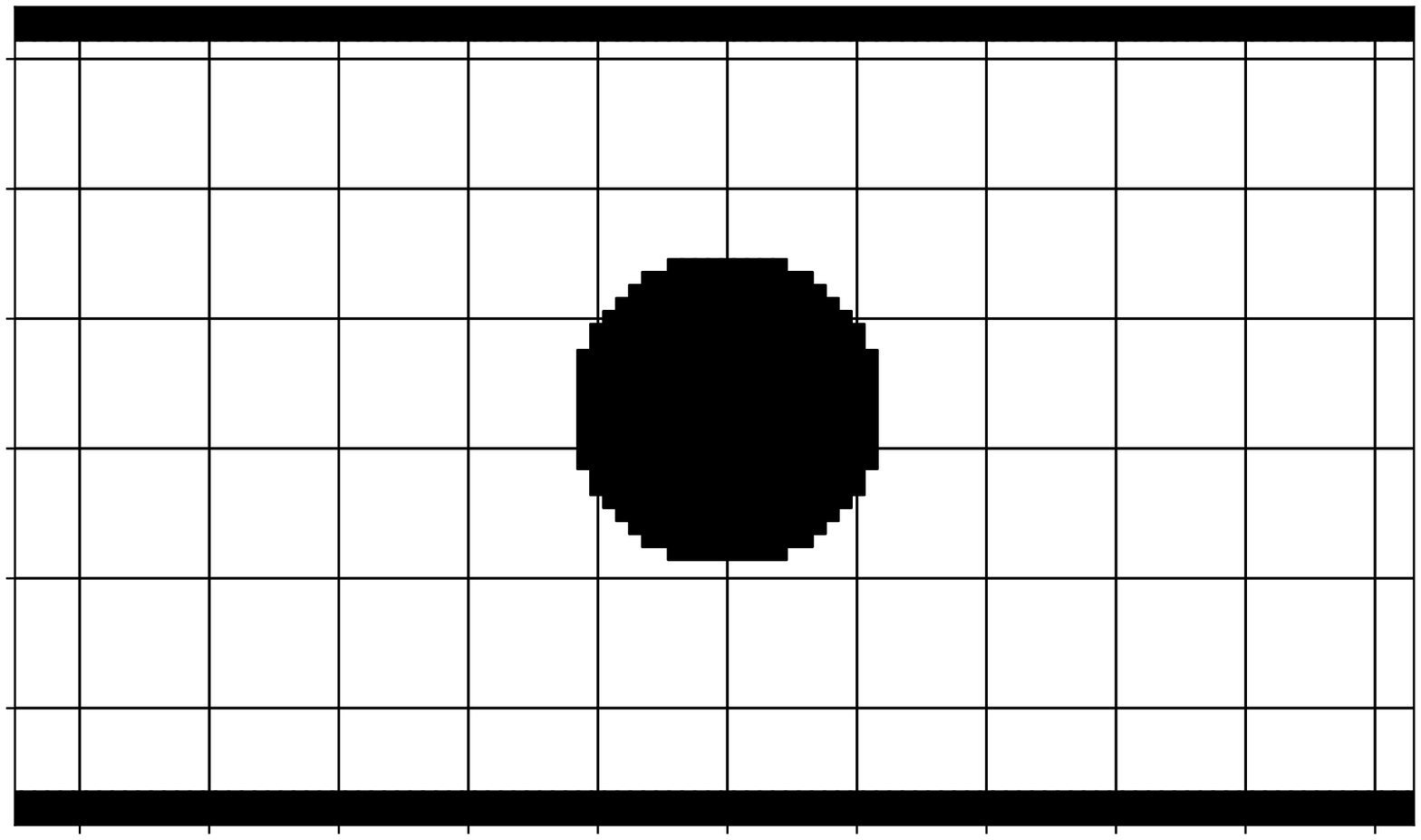} & 
		\includegraphics[width=20mm]{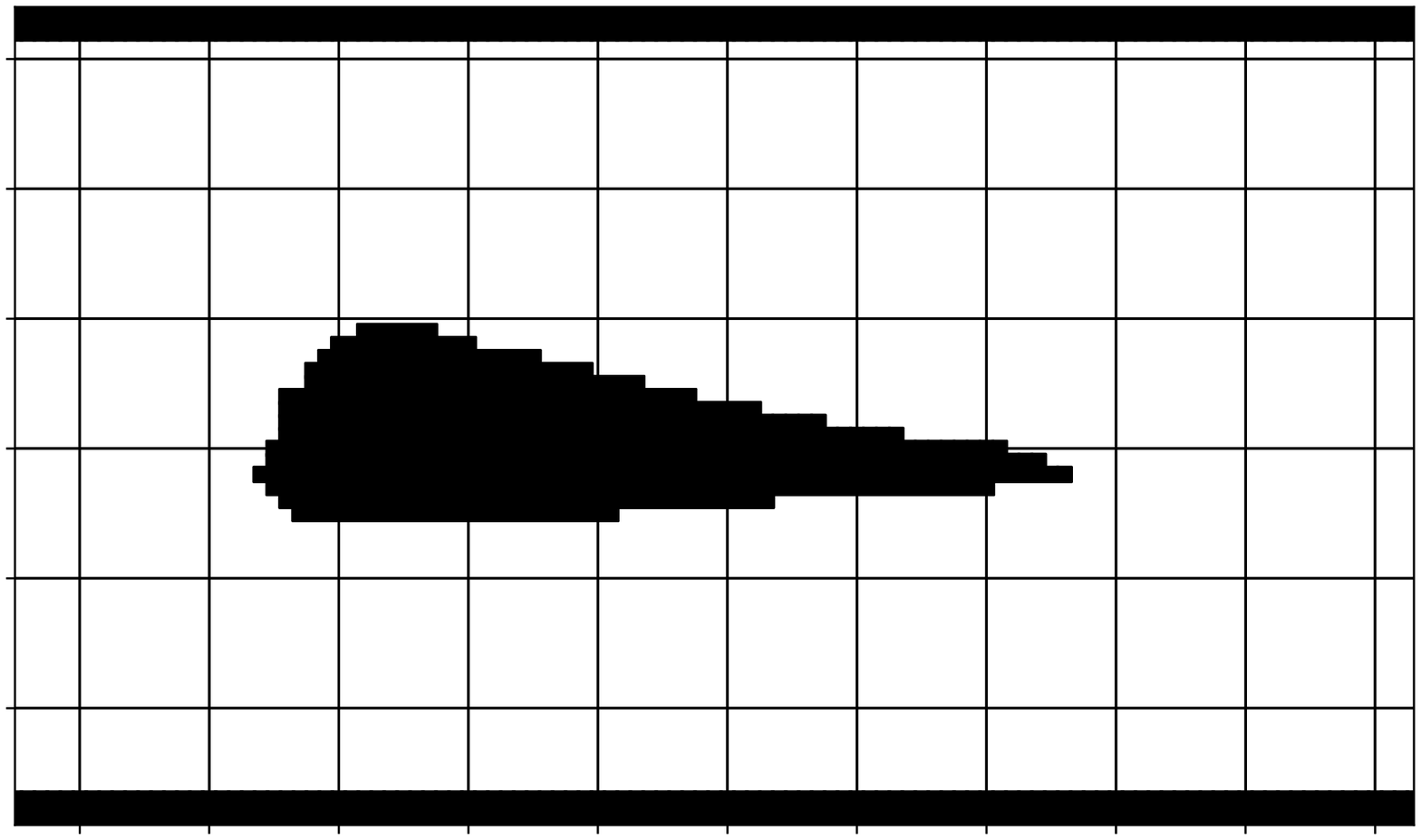} &
		\includegraphics[width=20mm]{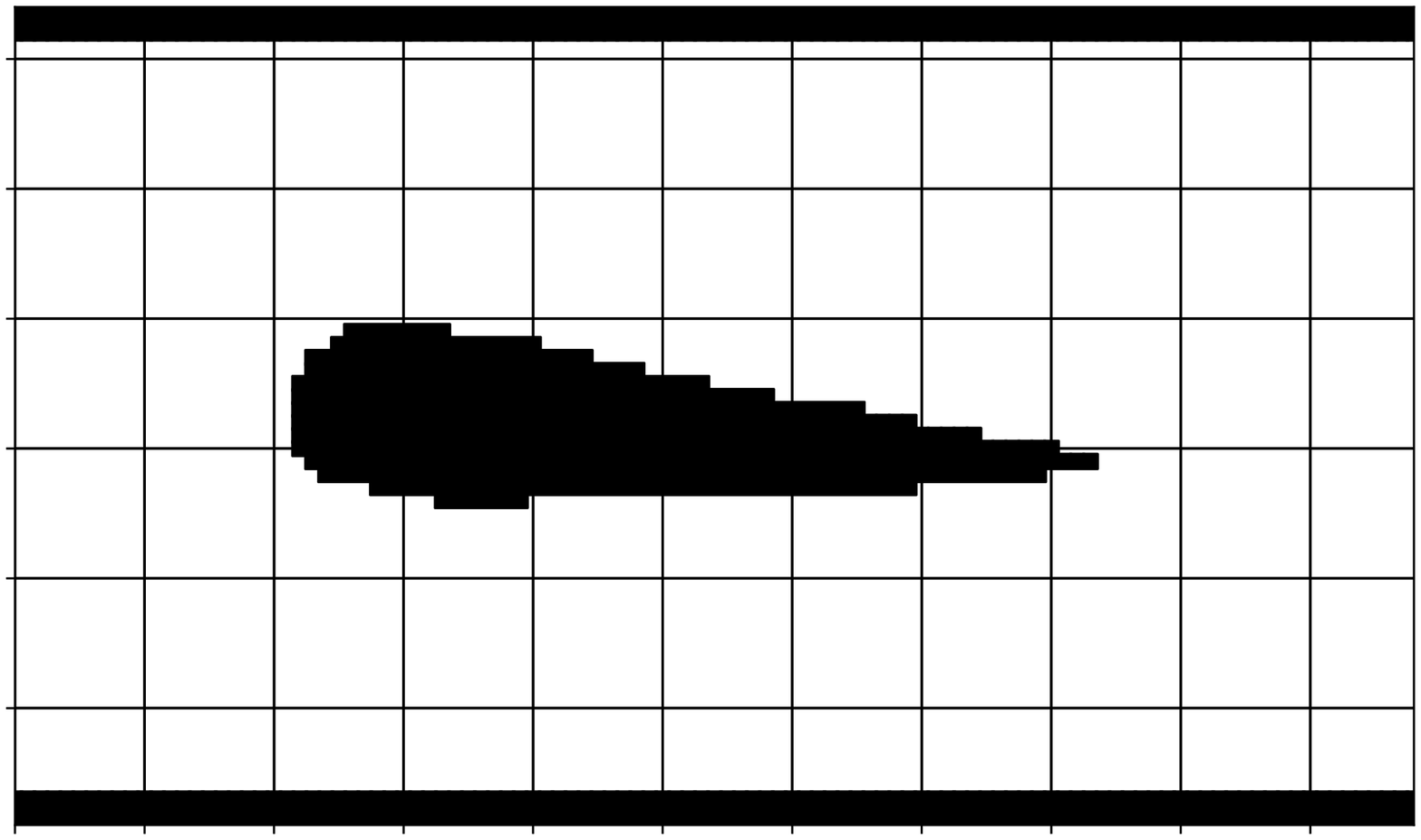} &
		\includegraphics[width=20mm]{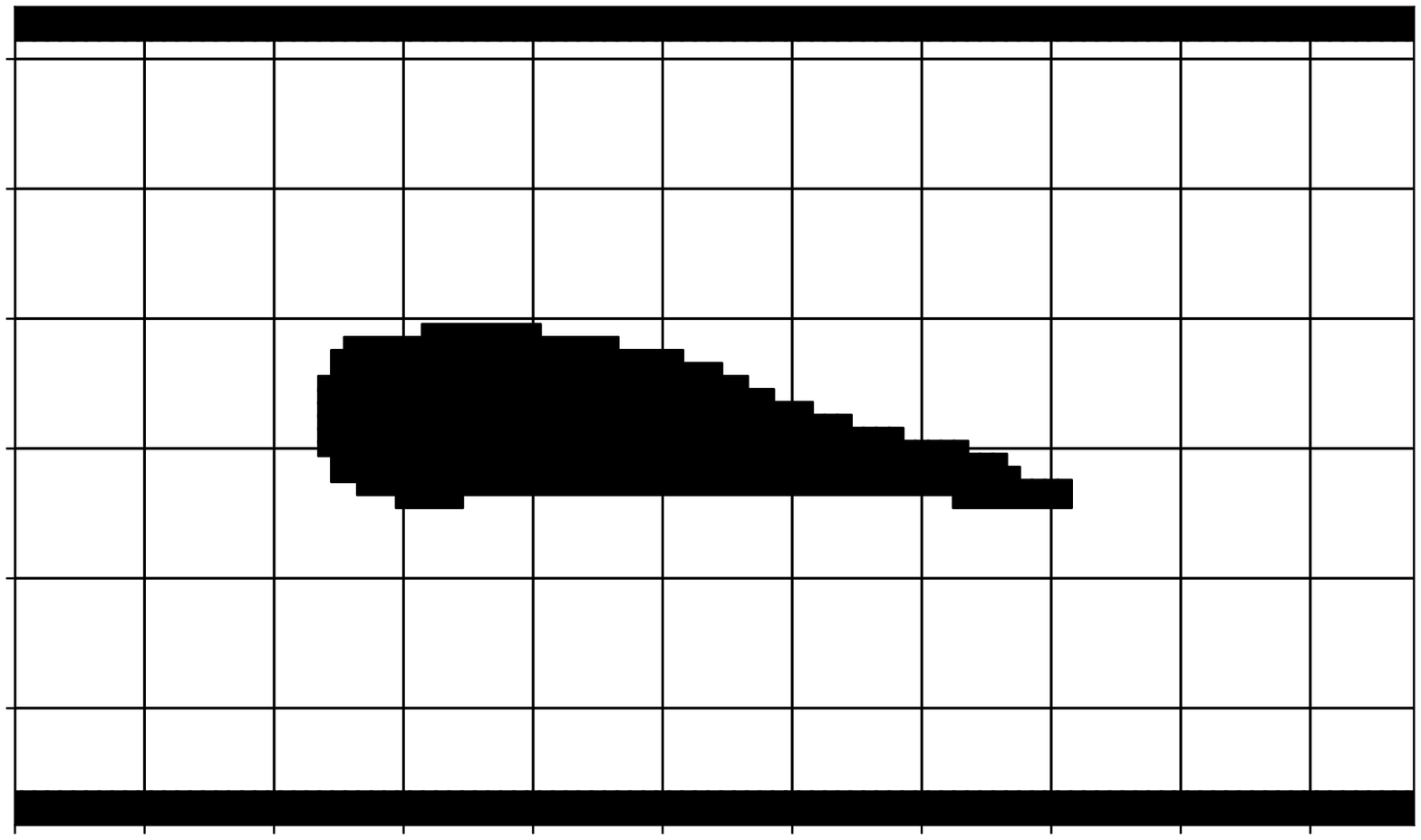} &
		\includegraphics[width=20mm]{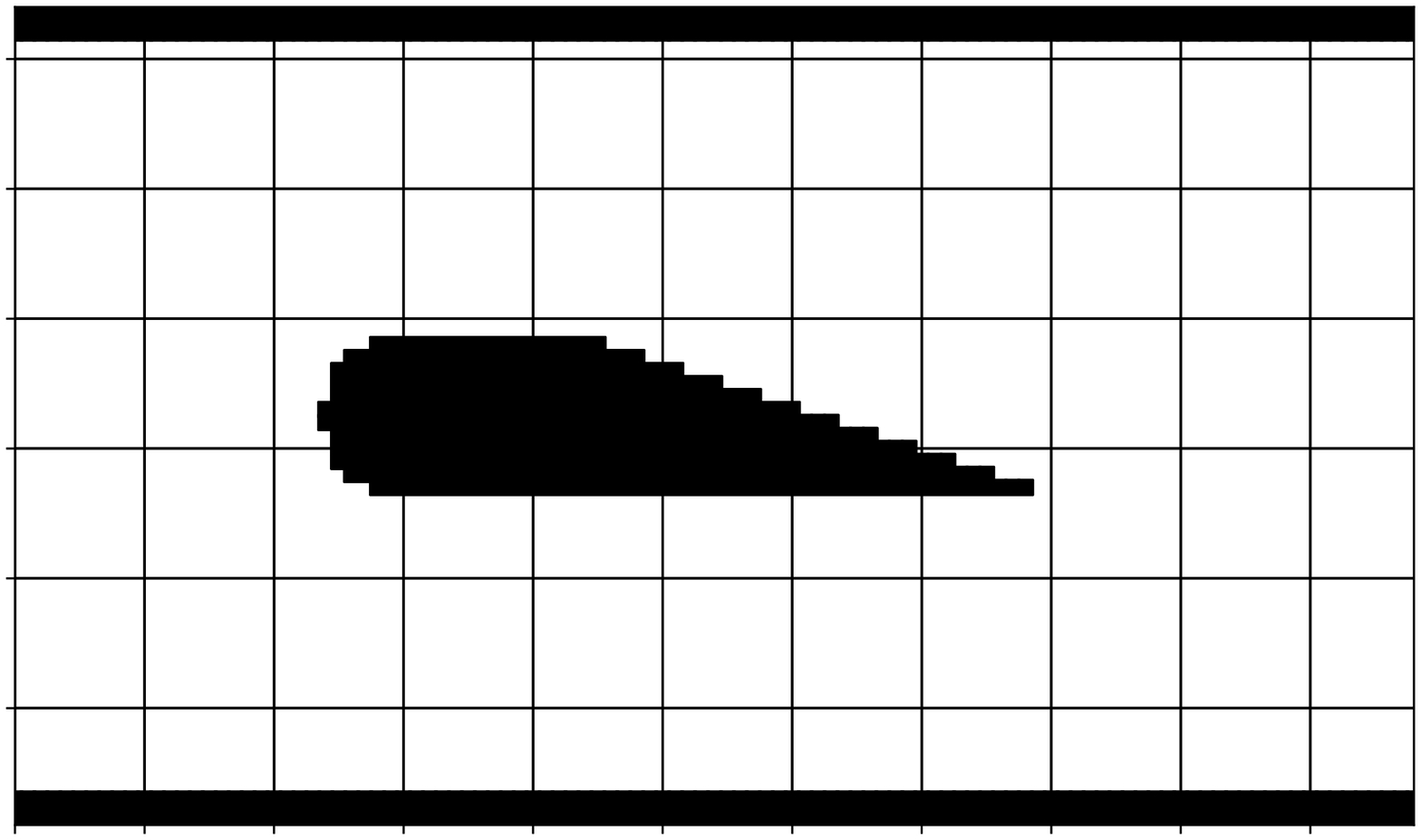}\\
		\includegraphics[width=20mm]{img/geos_transformed_sharp/circle.eps} & 
		\includegraphics[width=20mm]{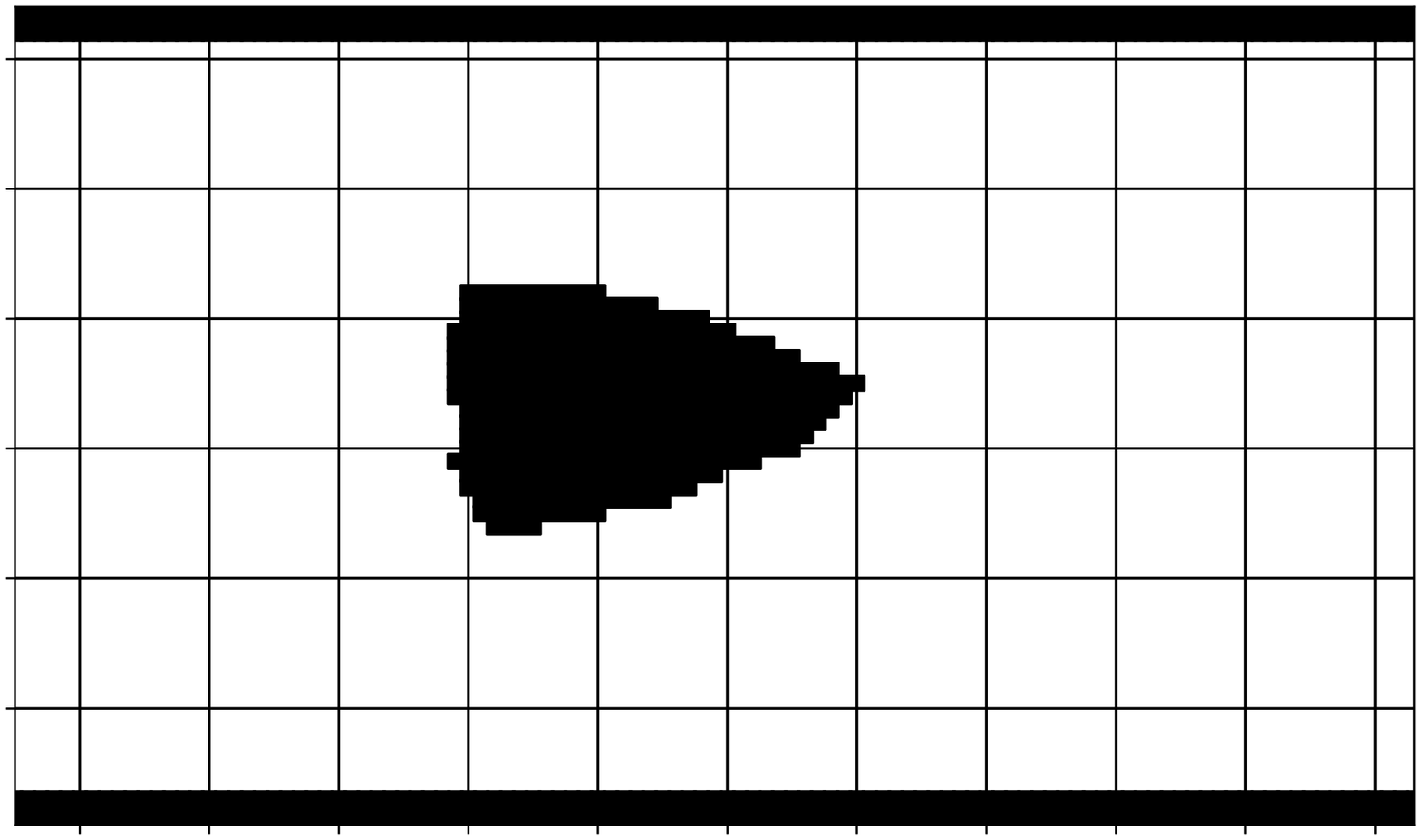} &
		\includegraphics[width=20mm]{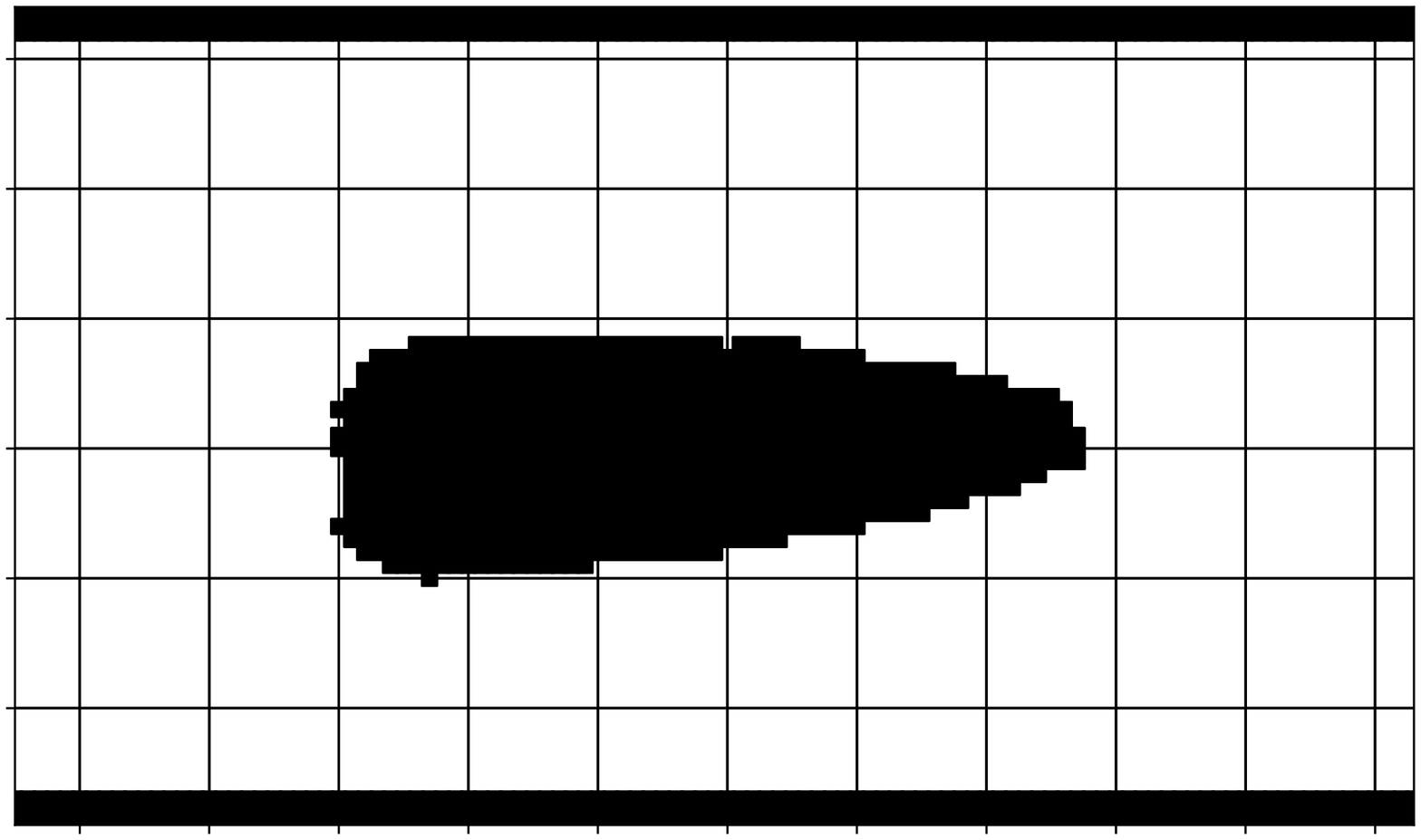} &
		\includegraphics[width=20mm]{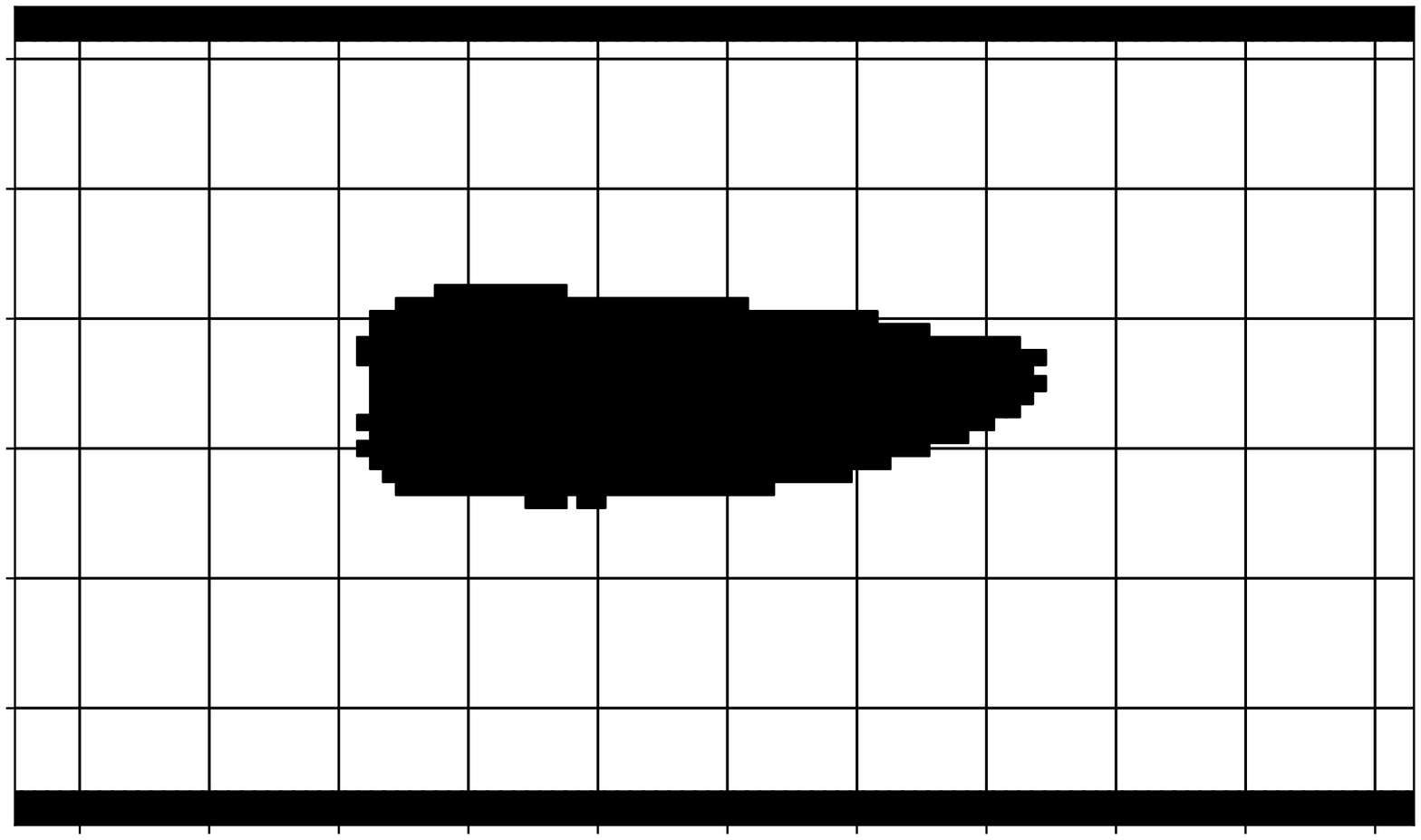} &
		\includegraphics[width=20mm]{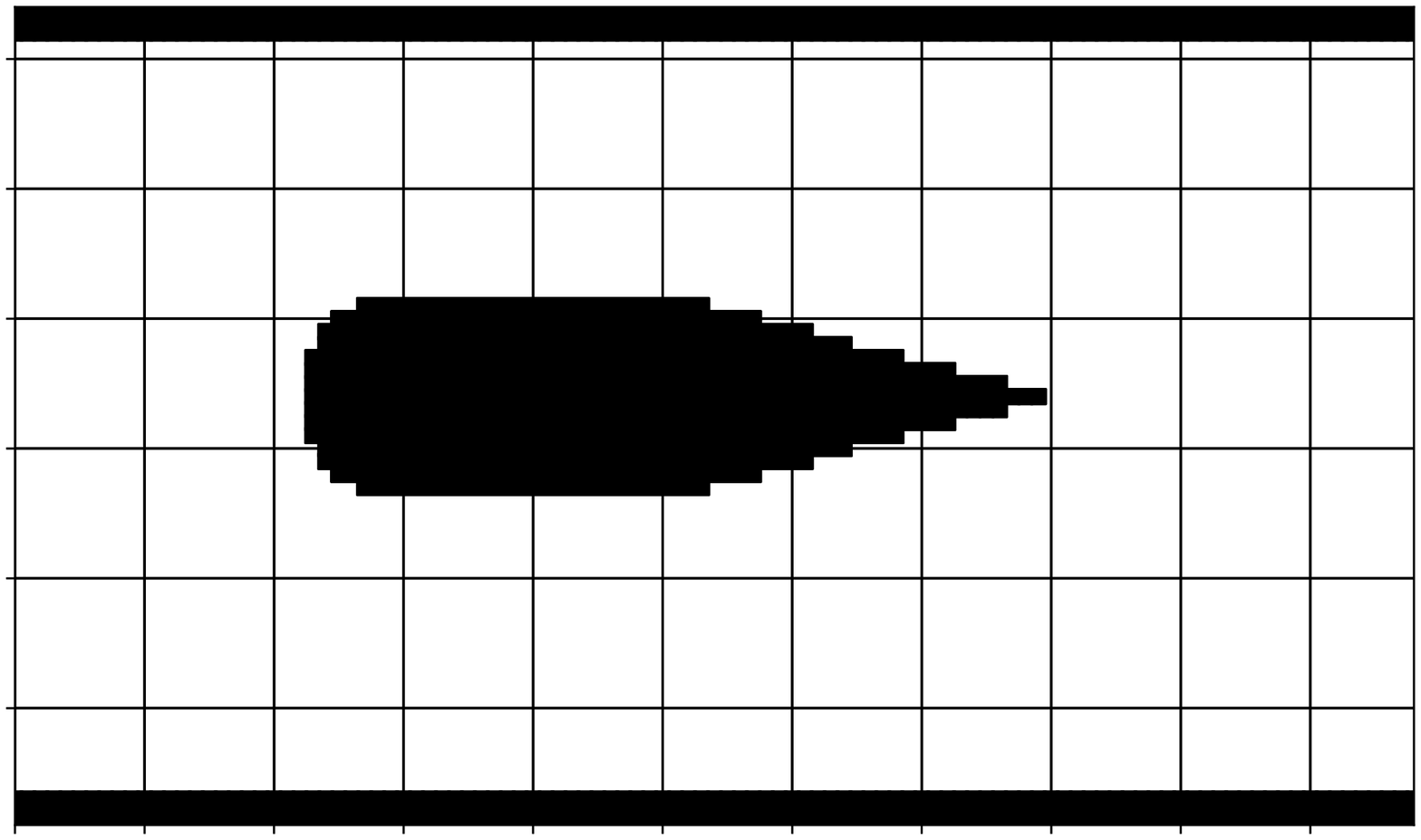}\\
		\includegraphics[width=20mm]{img/geos_transformed_sharp/circle.eps} & 
		\includegraphics[width=20mm]{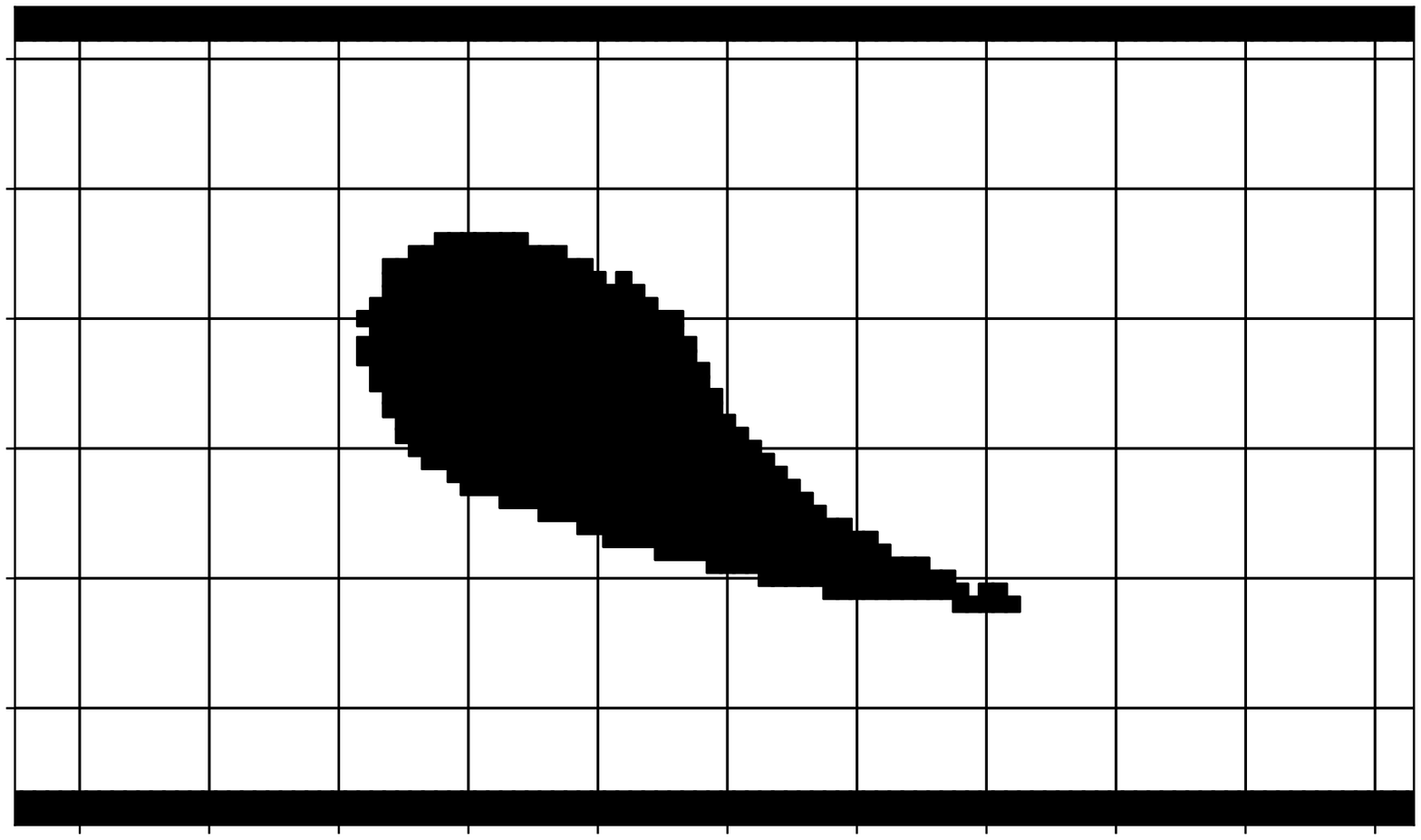} &
		\includegraphics[width=20mm]{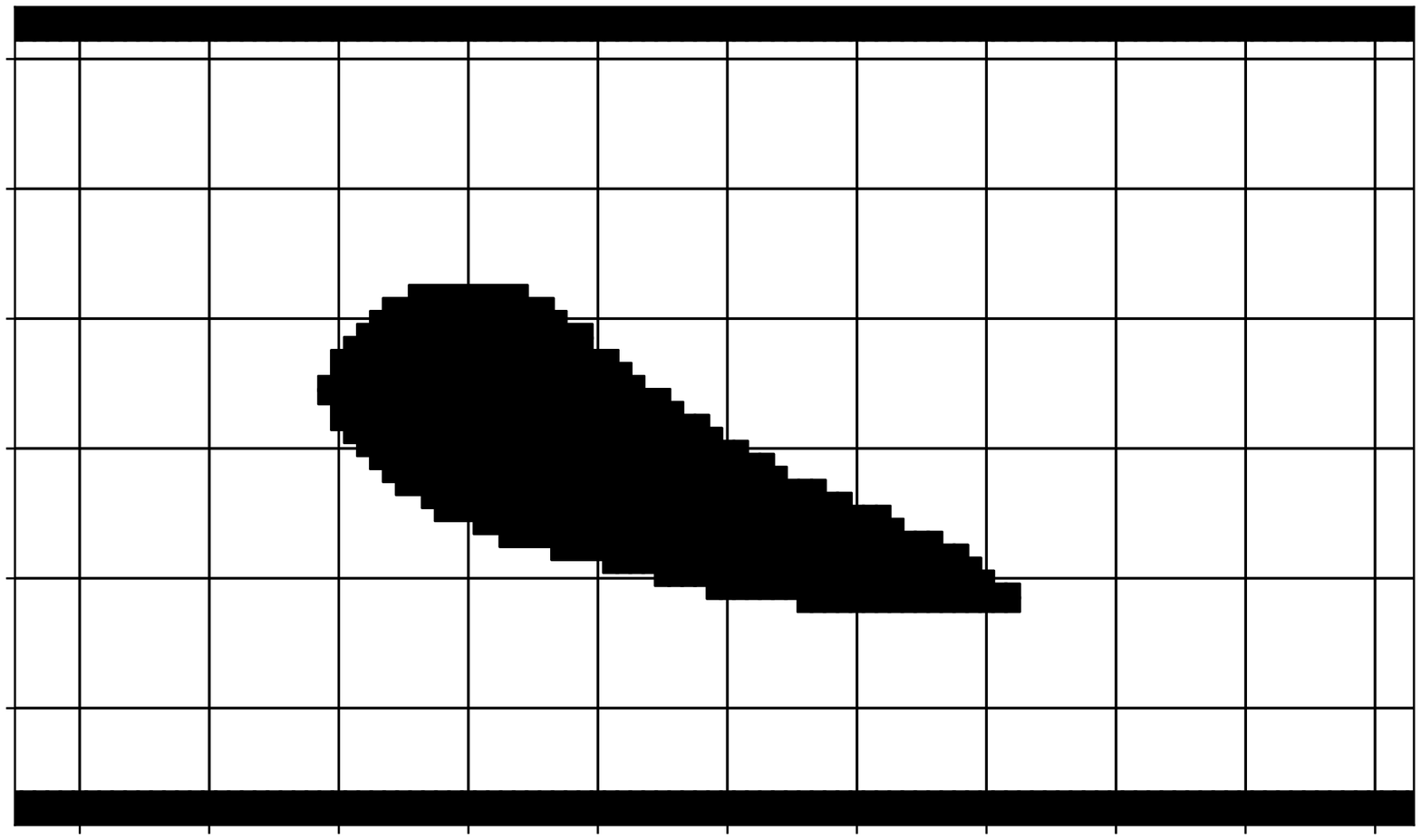} &
		\includegraphics[width=20mm]{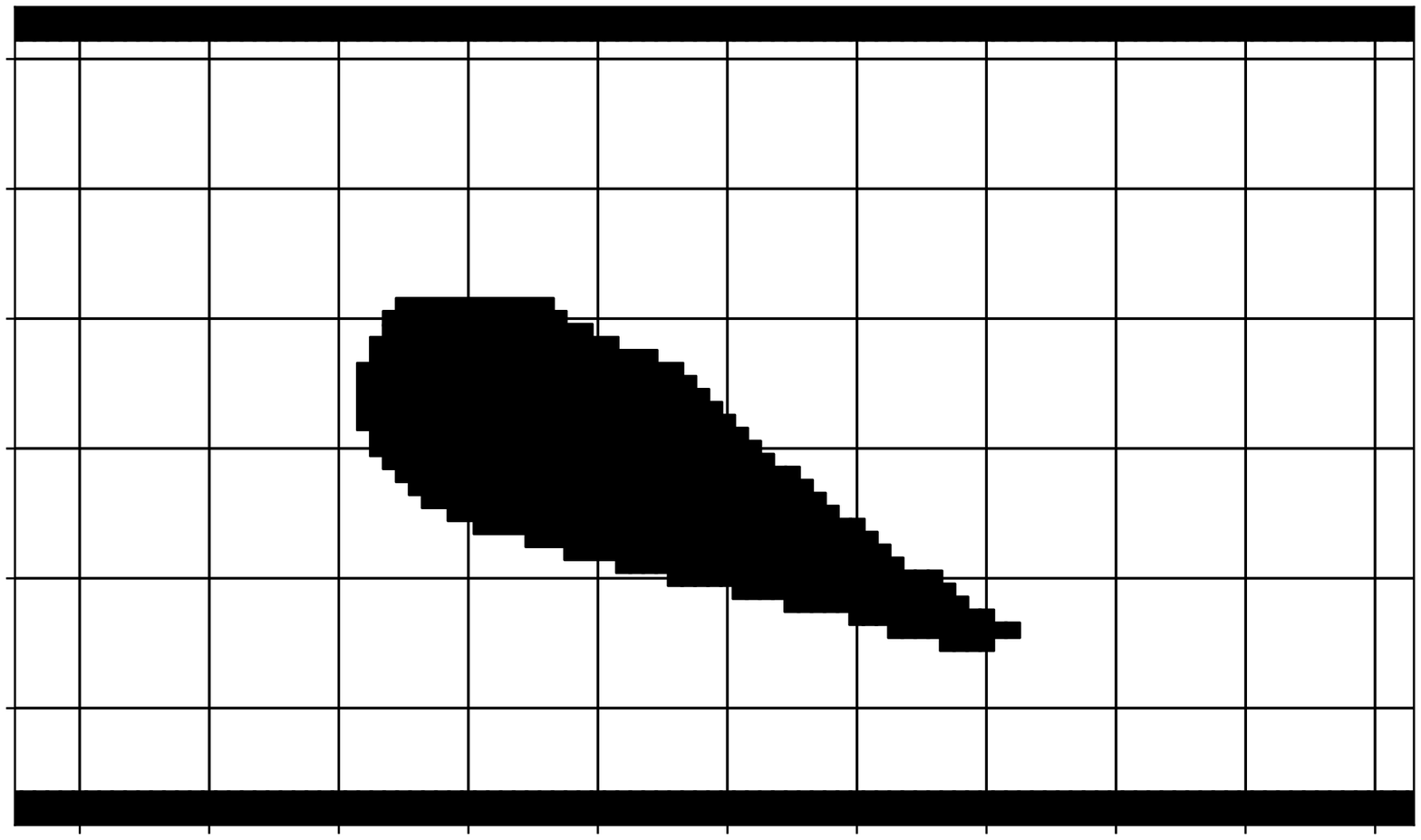} &
		\includegraphics[width=20mm]{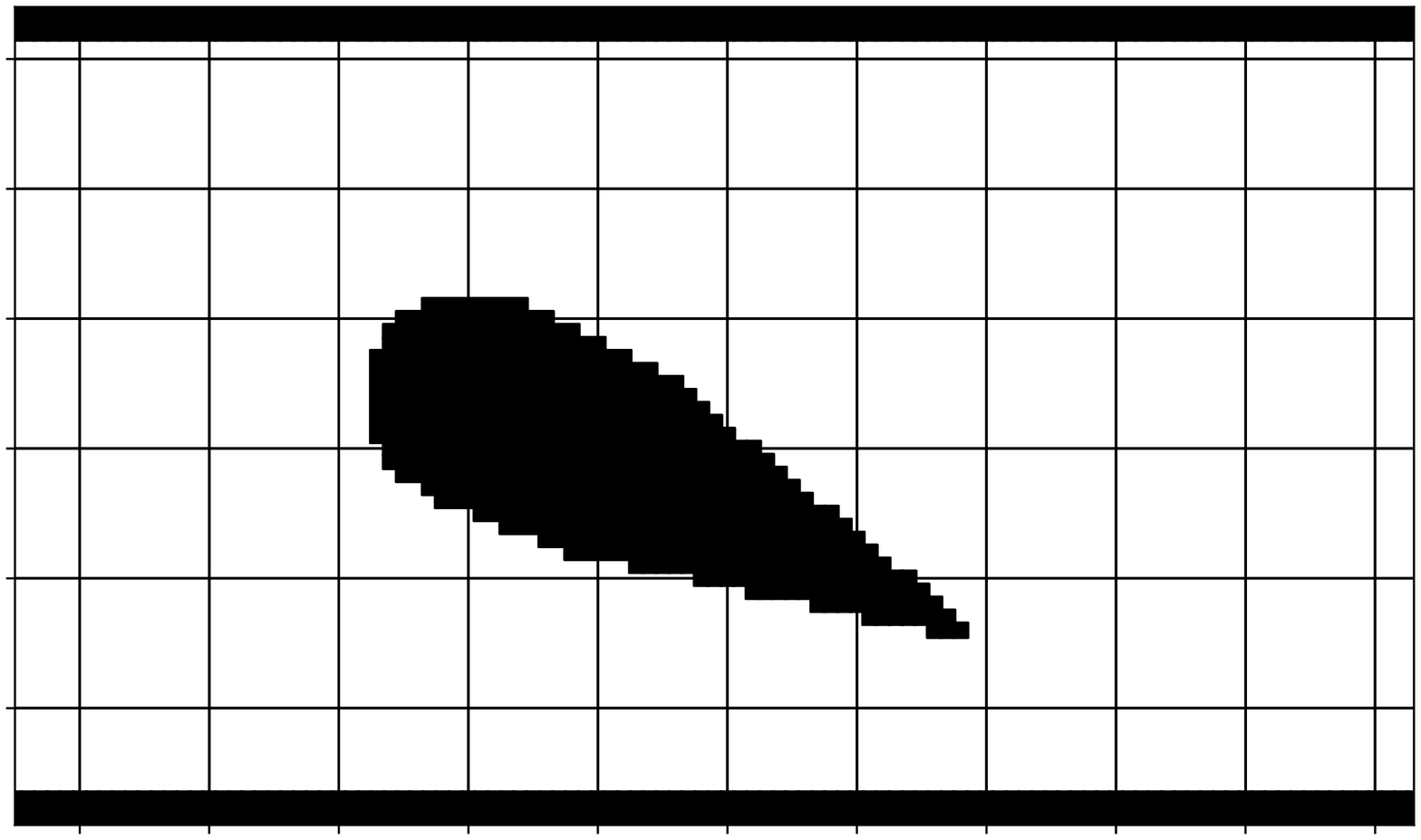}\\
		\includegraphics[width=20mm]{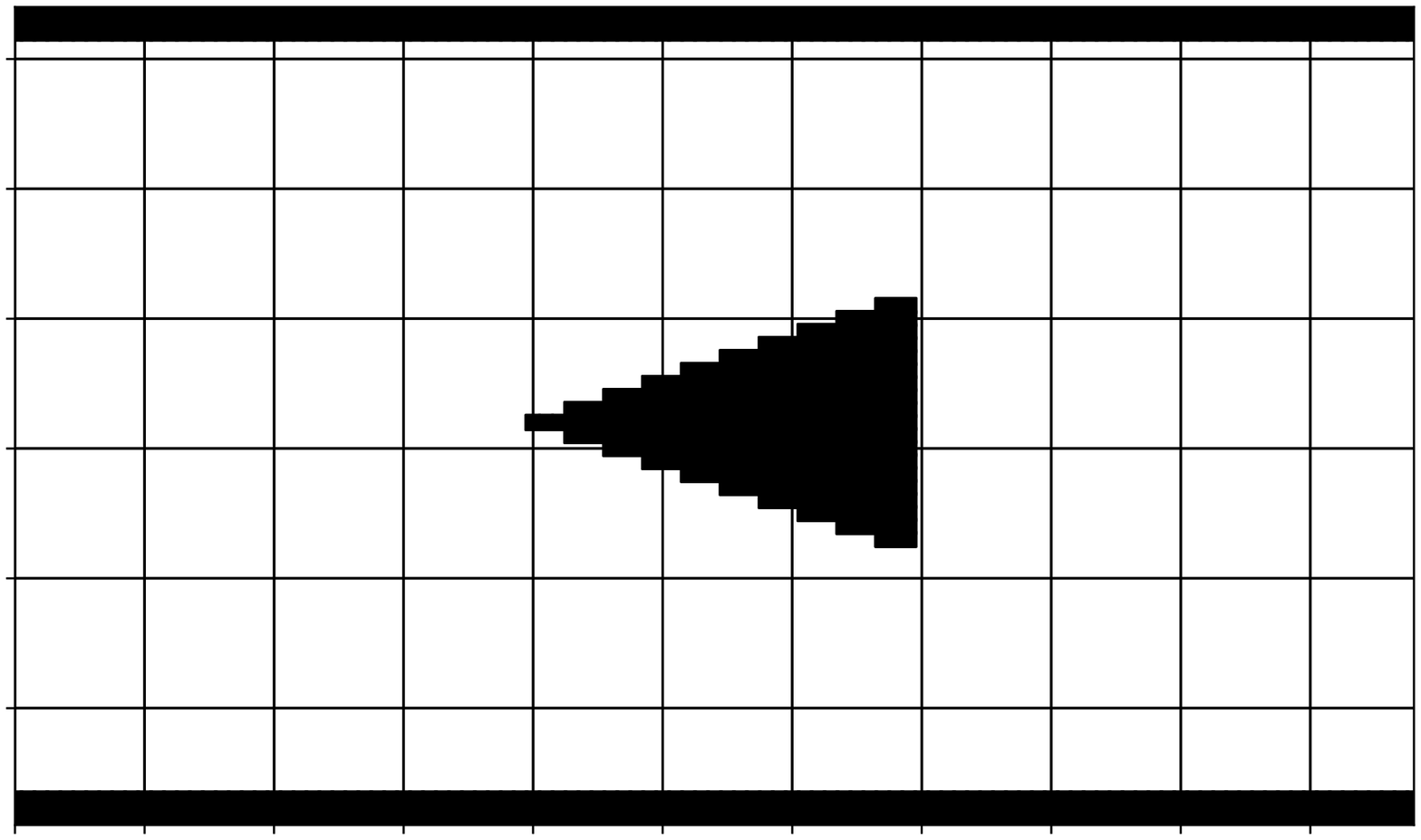} & 
		\includegraphics[width=20mm]{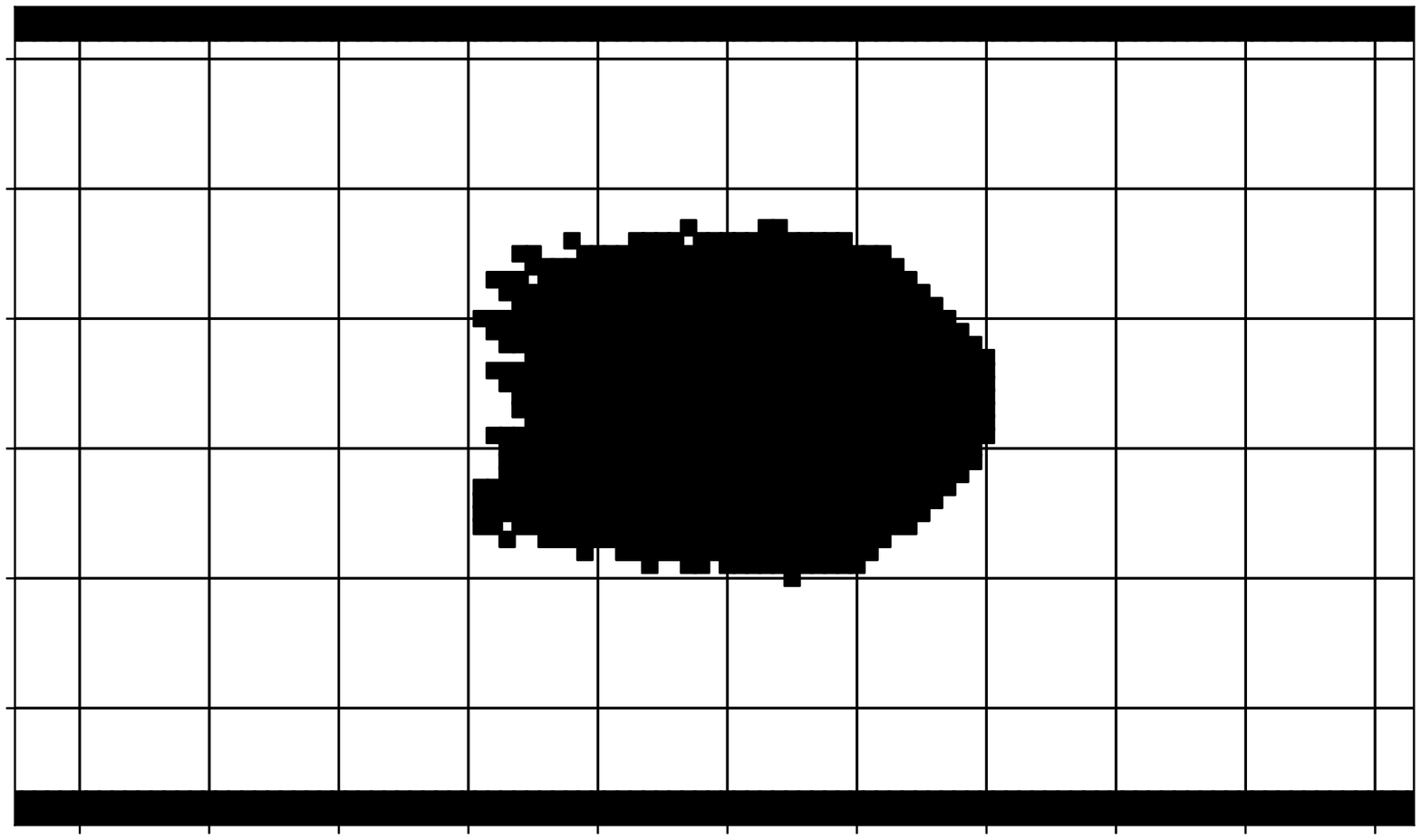} &
		\includegraphics[width=20mm]{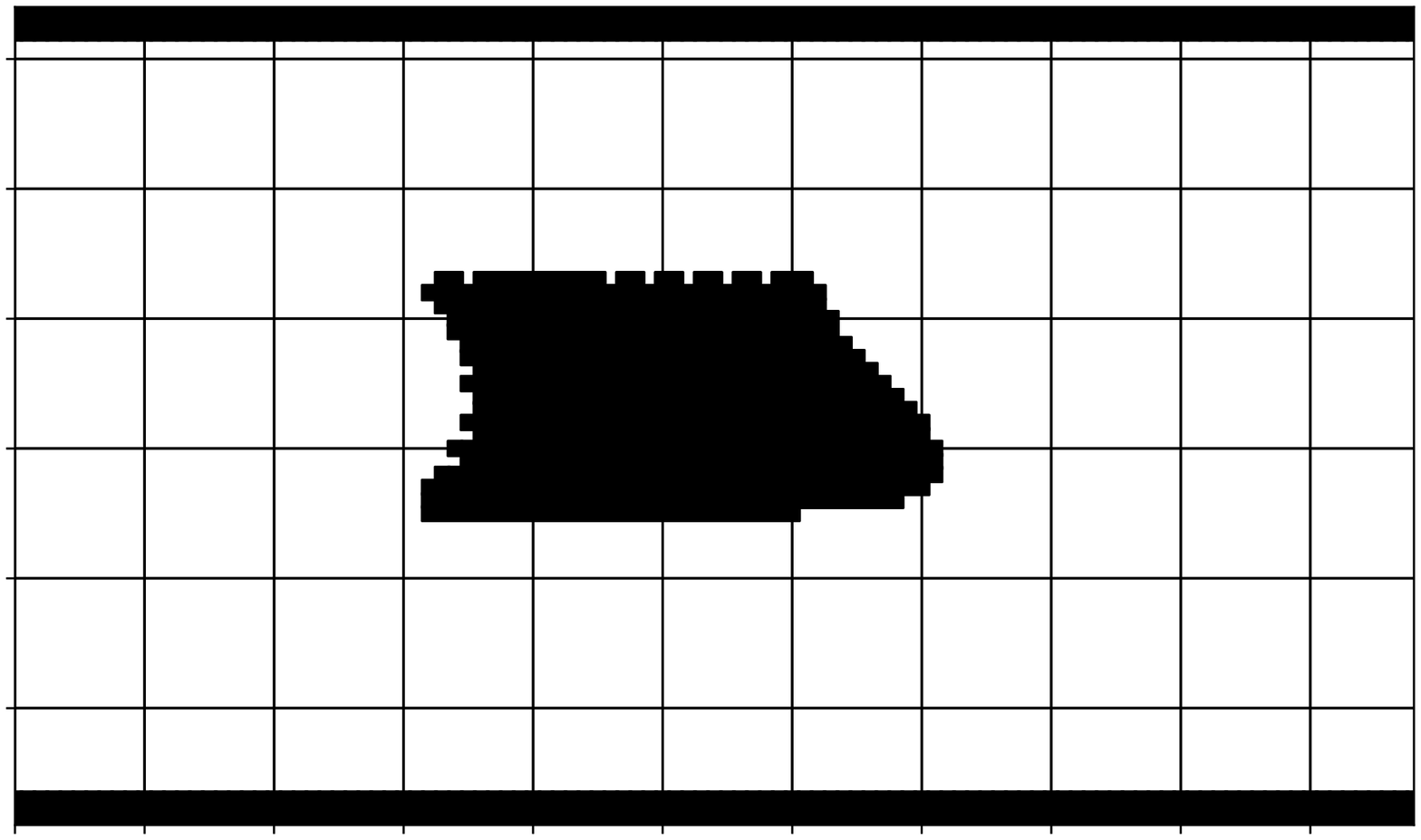} &
		\includegraphics[width=20mm]{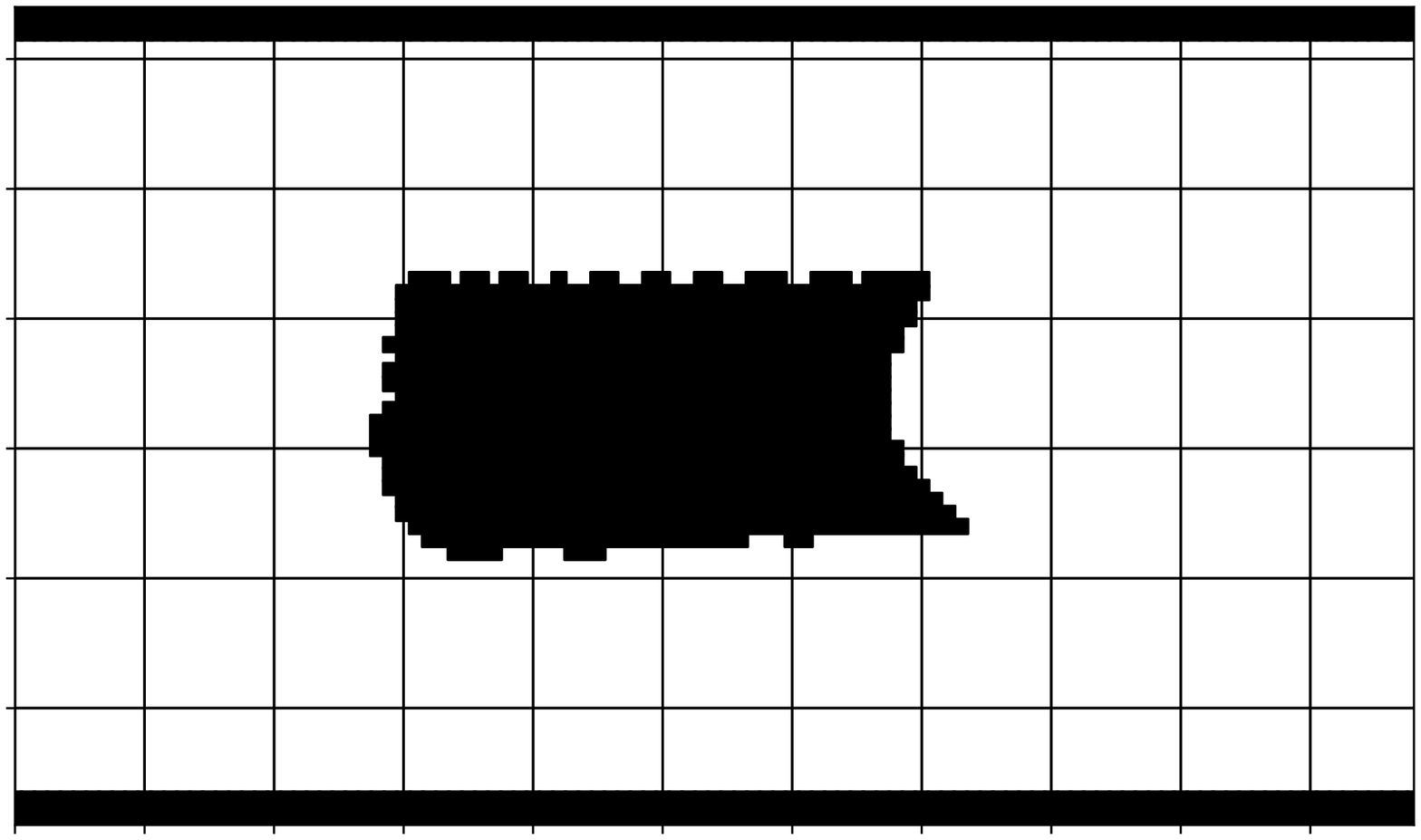} &
		\includegraphics[width=20mm]{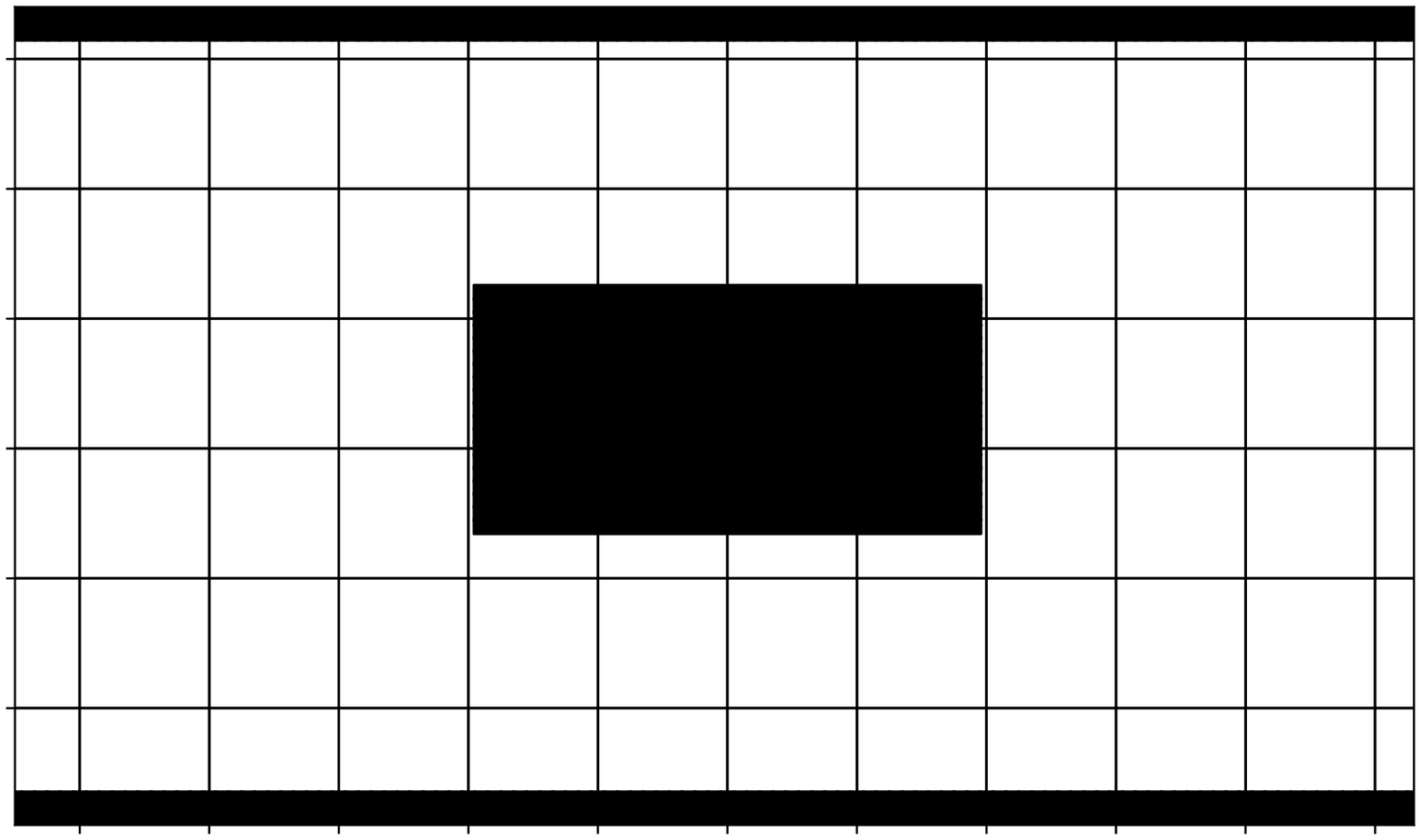}\\
		\includegraphics[width=20mm]{img/geos_transformed_sharp/rect.eps} & 
		\includegraphics[width=20mm]{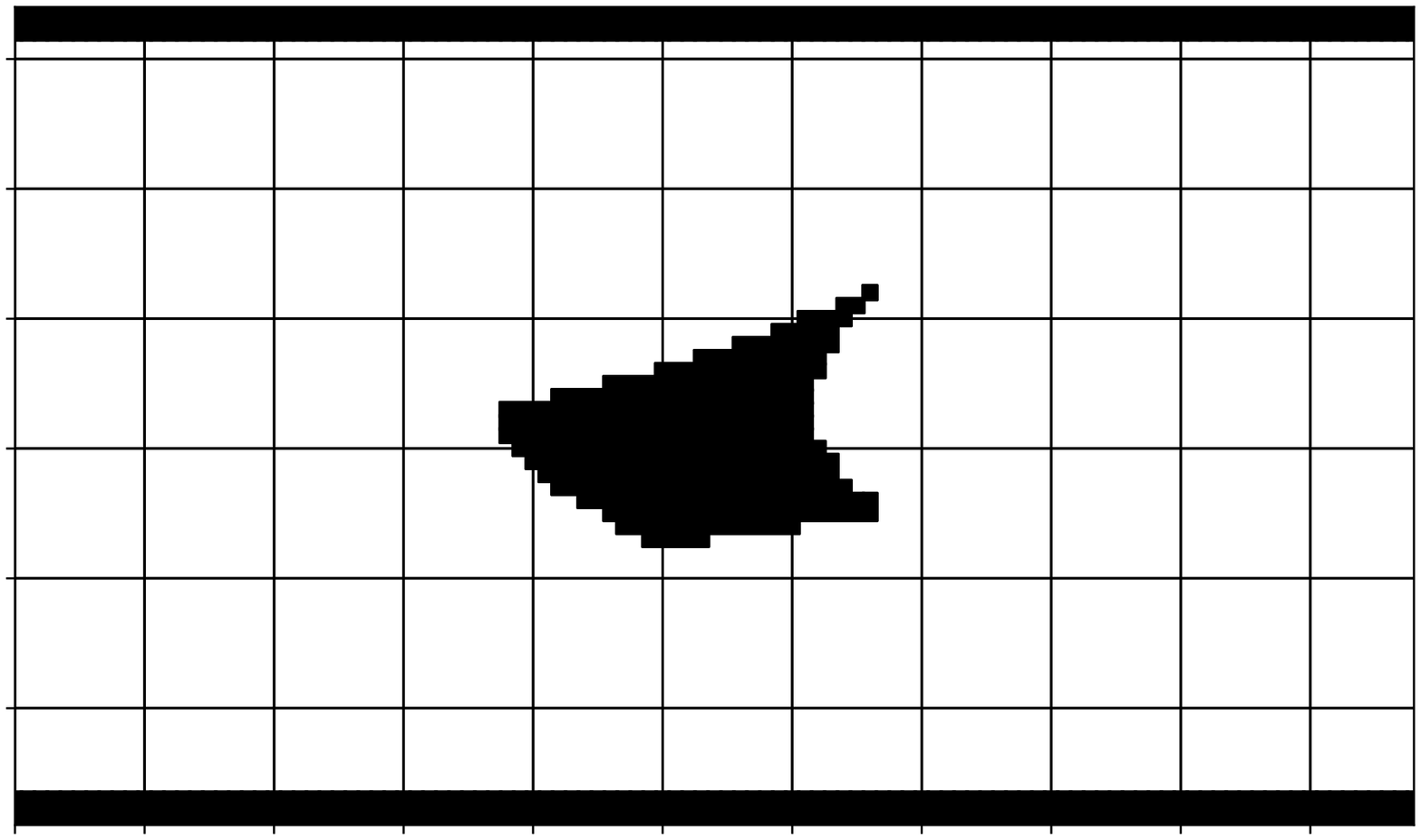} &
		\includegraphics[width=20mm]{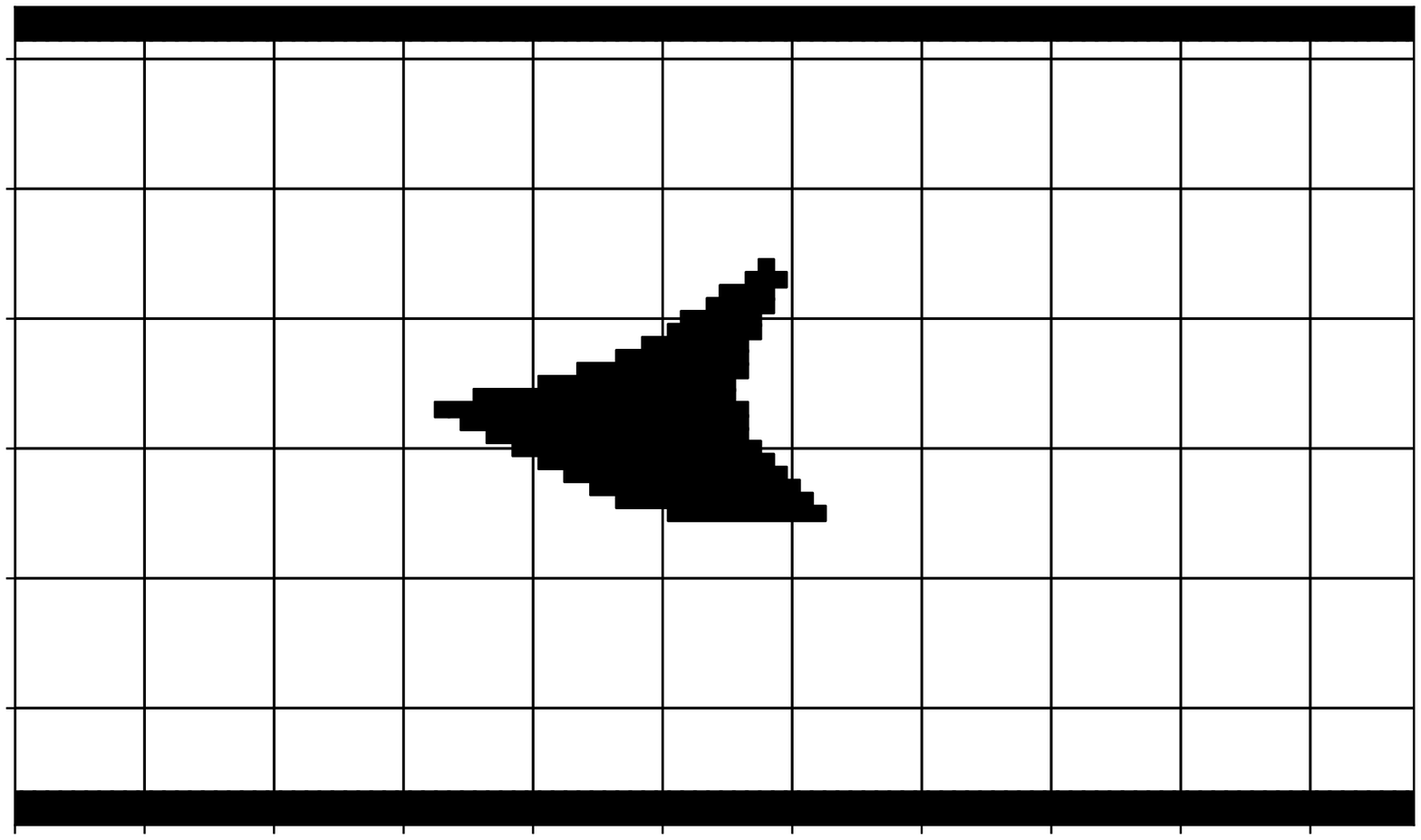} &
		\includegraphics[width=20mm]{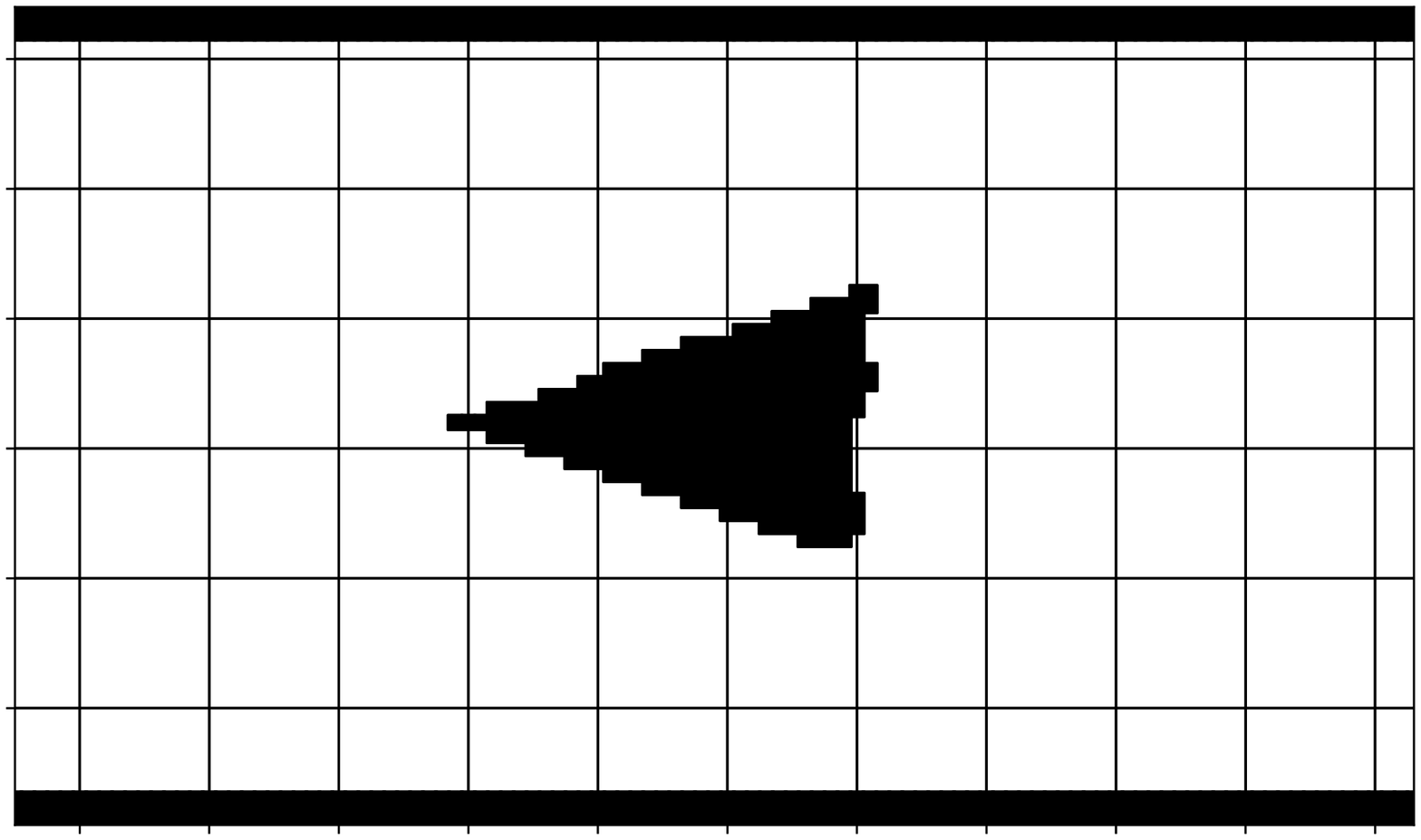} &
		\includegraphics[width=20mm]{img/geos_transformed_sharp/tri.eps}\\
		(a) Initial & (b) $\Delta p$, $F_i$ & (c) $\Delta p$, $F_i$, $L$ & (d) SSE & (e) Target\\
		& & \;\;\;\; $H$, $COM$ & &
	\end{tabular}
	%
	%
	\vspace{-0.2\baselineskip}
	\caption{Results of the shape transformation for the test cases with different constraints. (a) Initial geometry. (b) Total pressure difference $\Delta p$ and object forces $F_i$ constrained. (c) $\Delta p$ and $F_i$ with additional geometric restrictions. (d) All flow values $u$, $v$ and $p$ via $\mathcal{L}_{SSE}$. (e) Target geometry.}
	\label{resultsTrafo}
\end{figure*}
\section{Results}
To show the strong performance of the proposed method for the flow field prediction we use a dataset which has a high variance in terms of fluid mechanical properties and is more complex in comparison to those presented in related works. We show, that the shape optimization can be successfully applied to test cases with generic objects like rectangles or triangles. Further, we extend the experiments to real applications, such as the design of different airfoils. The design is optimized based on well known physical quantities which are typically used to analyse airfoils and are of relevance for the design of fluid machinery. Even for real applications we can achieve a good performance while keeping the flexibility of our model. Thus, we outperform other existing inverse design systems.
\subsection{Predicted Flow Fields}
\begin{table}[t]
	\center
	\caption{Relative error of $\Delta p$ and $F_i$ for the test cases with different constraints.}
	\label{resultsTable}
	\begin{tabular}{lccc}
		\hline
		\multicolumn{1}{c}{Target} & $\Delta p, F_i$ & $\Delta p, F_i, L, H, COM$ & SSE \\
		\hline\hline
		Airfoil (asym.) & 0.4\% & 0.01\%& 0.8\%\\ 
		\hline
		Airfoil (sym.) & 11.9\% & 1.1\% & 2.6\%\\ 
		\hline
		Airfoil (adjust.) & 0.7\% & 0.11\%& 0.14\%\\
		\hline
		Rectangle & 3.3\% & 0.06\% & 0.75\%\\
		\hline
		Triangle & 131\%& 0.8\% & 1.56\%\\
		\hline
	\end{tabular}
\end{table}
Since the shape transformation relies on the predicted flow field, the prediction of these values is a major issue in this method and a high accuracy is required. In this section a detailed analysis of the predicted flow fields is presented. We have 300 CFD-simulated flow fields as ground truth for the evaluation, which were not seen during training. First the accuracy over the whole flow field is presented. The mean relative error (MRE) of the normalized quantities is used for evaluation, since it yields an error percentage for the predicted solution. The reference value for the velocity channels~$u,v$ is the averaged inflow velocity. For the pressure channel~$p$ the averaged pressure at the outlet is used. These reference values are chosen since the inlet velocities and the outlet pressure are also used as boundary conditions for the CFD-simulation. The errors are not equally distributed over the three channels, which yield relative errors of 4.34\% for the $x$-velocity channel, 2.5\% for the $y$-velocity and 1.88\% for the pressure. Especially the pressure values have a small error, which is useful since the pressure forces~$F_i$ used for the shape optimization only depend on the pressure values.\\
\indent
Further the distribution of the MRE in $u,v$~and~$p$ along the channel length is analysed. The channel length is represented by 256 pixels along the $x$-direction and the MRE is calculated per column of the associated channel of the data matrix. The results are shown in Figure~\ref{relativeErrors} (left). The velocities in the middle part of the channel and downstream are varying more than the velocities at the inlet because of the objects, which are diverting the flow. This results in increasing errors of the predicted velocities along the channel length. In contrast, stagnation points in front of the objects result in larger pressure gradients in the front part of the channel and thus in decreasing errors of the predicted pressure values over the length.\\
\indent
Since the interest is in physical quantities calculated on the flow field, the continuity equation $\Delta Q$ and the total pressure difference $\Delta p$ are analysed. The continuity equation is given as
\begin{equation}
 \Delta Q = \rho u_2 A_2 - \rho u_1 A_1\;,
\end{equation}
with density~$\rho$ and $A_{1,2}$ denoting the inlet and outlet surface, respectively. The velocity $u$ is again averaged over the cross-sectional area of the channel. The inlet position is fixed to the first column of the data matrix. The outlet position is shifted stepwise over the channel length. The MRE is also used here to analyse $\Delta Q$ and $\Delta p$ normalized to the mass flow \mbox{$Q_1=\rho u_1 A_1$} and the energy term $\big(\frac{\rho}{2} (|u_1+v_1|^2)+p_1\big)$ at the inlet, respectively. From Figure \ref{relativeErrors} (right) it is visible that the error of $\Delta Q$ is staying below 2\%. Also the error in $\Delta p$ is low for broad areas of the channel. This is important since $\Delta p$ is further used in the loss function $\mathcal{L}_{total}$ for the shape optimization. 
\subsection{Transformed Geometries}
\begin{figure}[t]
\center
\begin{tabular}{cc}
	\includegraphics[width=40mm]{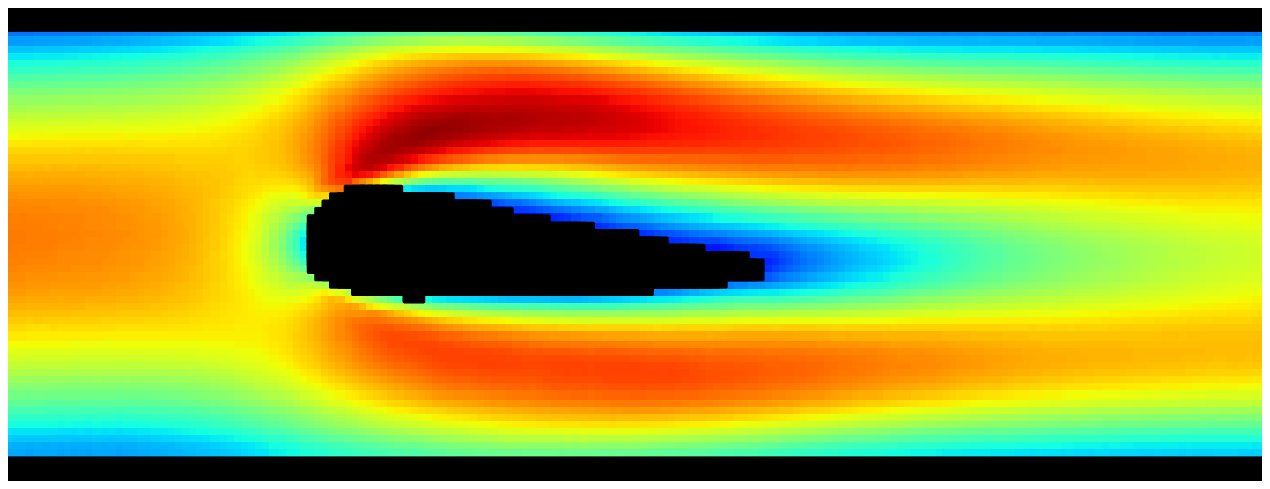}&\includegraphics[width=40mm]{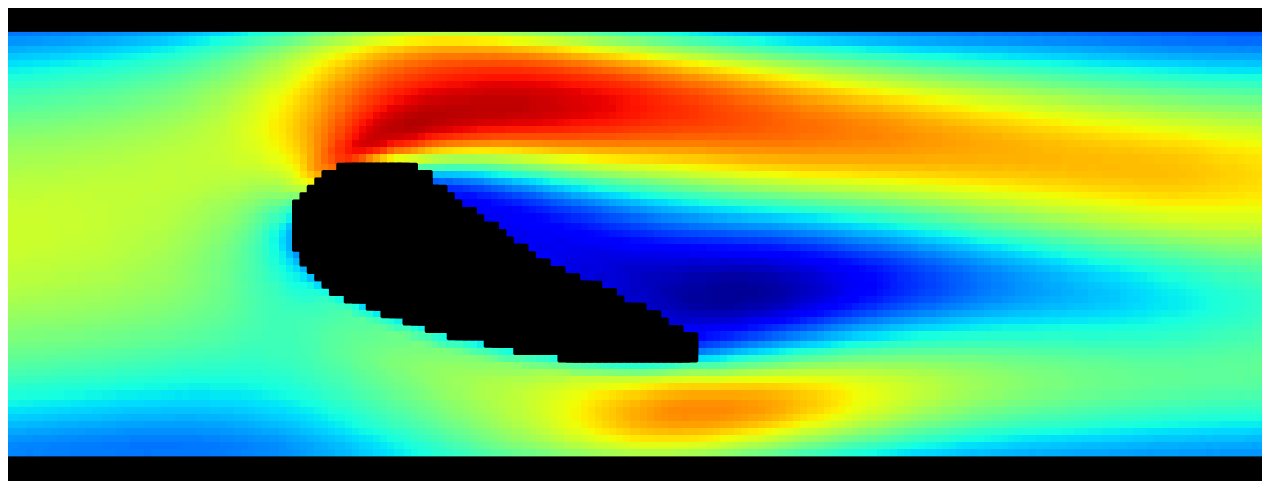}\\
	\includegraphics[width=40mm]{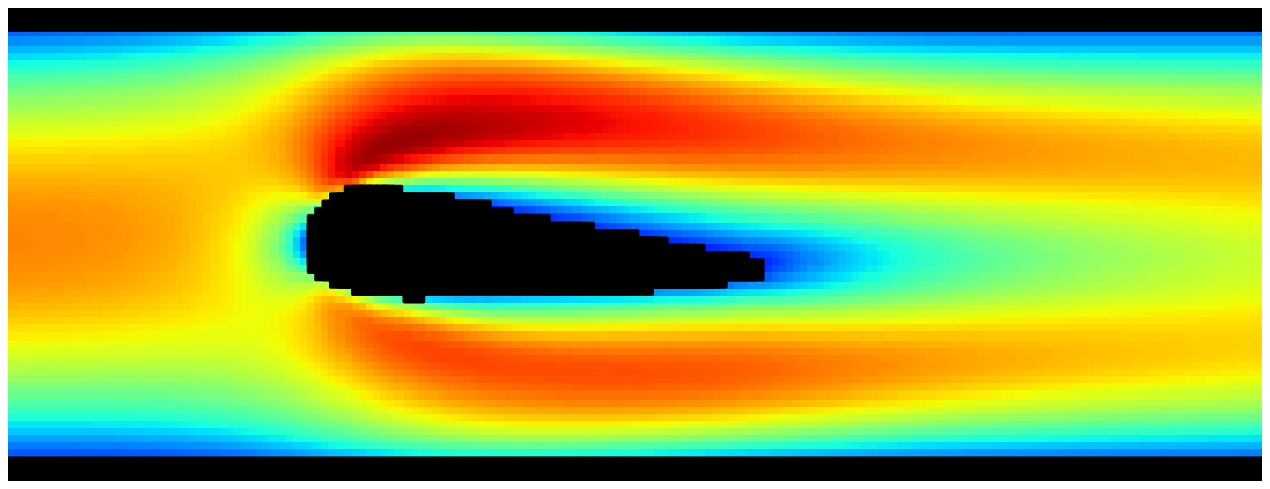}&\includegraphics[width=40mm]{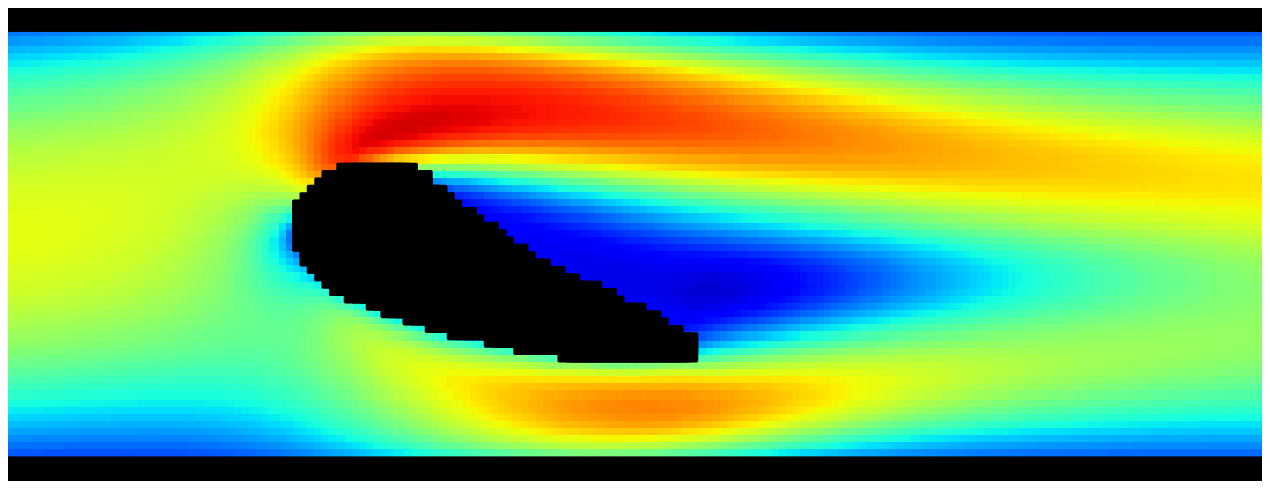}\\
	\includegraphics[width=40mm]{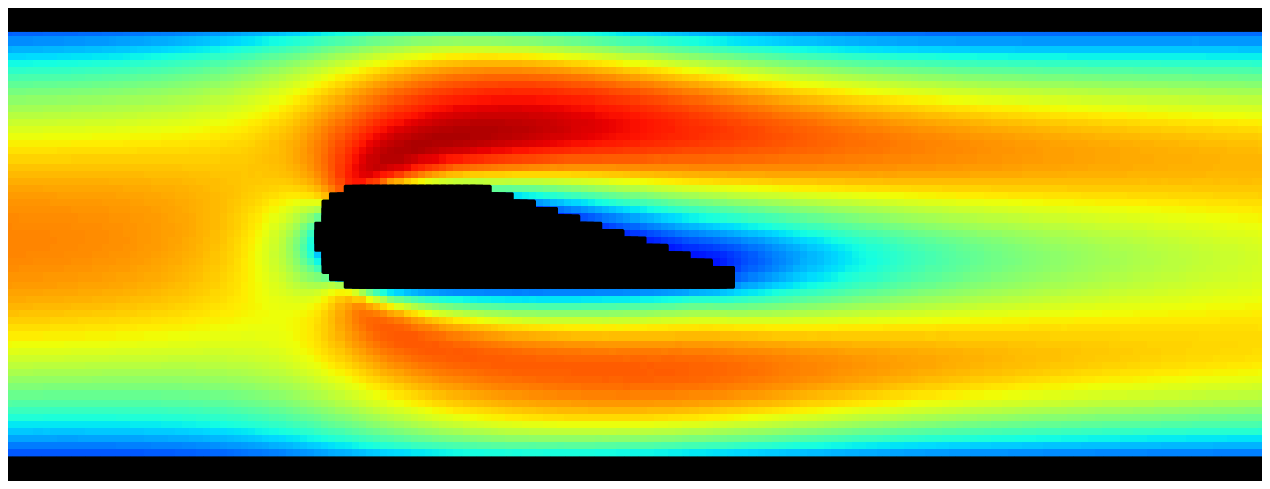}&\includegraphics[width=40mm]{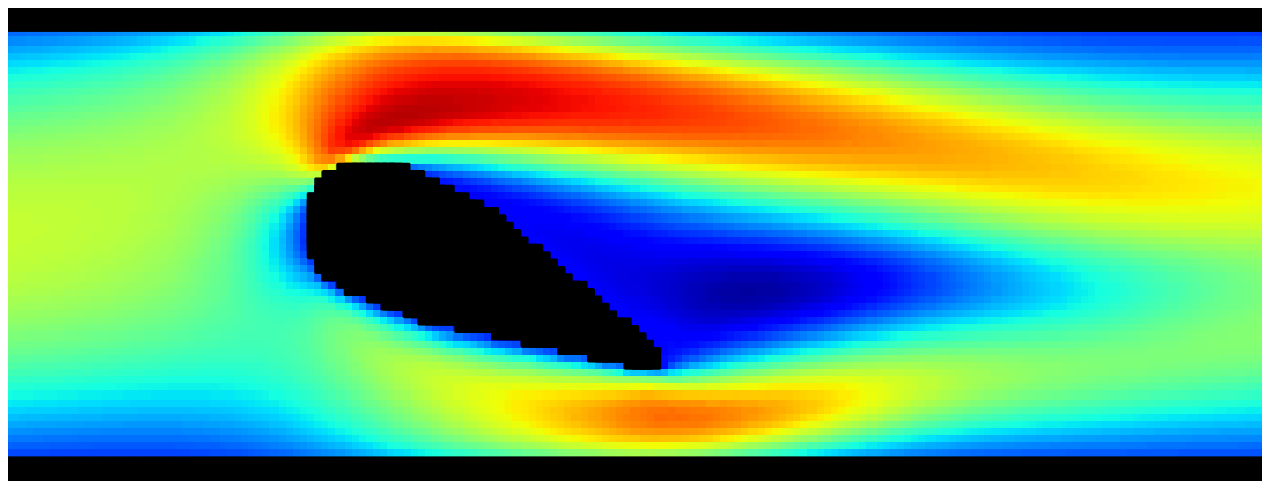}
\end{tabular}
\caption{Validation of the transformed geometry. \textit{Top}: Predicted values $u$. \textit{Middle}: CFD-simulated values $u$. \textit{Bottom}: CFD-simulation of the target.}
\label{comparisonPredCFD}
\end{figure}
The previous shown results give a solid basis for the following shape transformation. We will present three configurations with different physical constraints. First the total pressure difference $\Delta p$ and the object forces $F_i$ will be used as constraints. Second, additional the object length $L$, height~$H$ and the centre of mass $COM$ will be restricted to match the target geometry. For the last configuration all values of~($u$,~$v$,~$p$) over the whole field will be used as constraints for the transformation via the $\mathcal{L}_{SSE}$. This configuration is used to show the maximum potential of the proposed transformation method.\\
\indent
%
For every configuration different initial and target geometries are used. This includes generic objects such as a circle, rectangle and triangle, and more realistic objects such as a symmetric and an asymmetric airfoil, and a airfoil with an adjusted angle. This leads to 15 different test cases. The columns (a) and (e) in Figure \ref{resultsTrafo} are showing the used initial geometries and the respective target geometries. The flow field of the target geometries is used to calculate the physical quantities which are used as target values for the geometry transformation, i.e. ground truth. 
Table \ref{resultsTable} shows the relative error between the ground truth and the physical quantities calculated on the estimated flow field resulting from the transformed geometry. For all the test cases only the mean relative error of $\Delta p$ and $F_i$ is shown for being able to compare the influence of the different constraints.\\
\indent
From Figure \ref{resultsTrafo} (b) it is visible that restricting $\Delta p$ and $F_i$ yields to transformed objects which are somehow looking like the target objects. Especially the asymmetric airfoil in the first row and the airfoil with adjusted angle in the third row have high similarities to their target objects. However, by using only $\Delta p$ and $F_i$ as constraints the solution space is broad. Thus, many objects with different shapes can fulfil these constraints. For example a symmetric airfoil as shown in the second row has equal forces on the top and bottom surface. It is visible that also the transformed airfoil has a symmetric shape. The relative error in Table \ref{resultsTable} is showing that the parameters of the transformed symmetric airfoil are close to these of the target airfoil. Only the triangle could not be transformed successfully. The sharp corner at the front of the triangle is leading to strong varying pressure distributions at this point even for small geometric variations. Thus, the system is not able to converge to a proper solution, since the loss function is dominated by the pressure distribution.\\
\indent
With additional constraints for the object length $L$, height~$H$ and the centre of mass $COM$ the solution space is getting small. This has a large impact to the transformation leading to results which are similar to the target shape. This is valid for all examples as shown in Figure \ref{resultsTrafo} (c). However the generic objects have more difficulties during the transformation, since the sharp corners are still causing strong varying pressure distributions. Also the relative error of $\Delta p$ and $F_i$ is decreasing for all examples to values smaller than $1.1\%$ which is showing the power of these physical constraints for transforming the objects.\\
\indent
To further analyse the maximum potential of the transformation all values of $u$, $v$ and $p$ over the whole field are used as constraints via the $\mathcal{L}_{SSE}$ which is indeed a strong restriction. Also the flow field of a target geometry is in general not known in a real world application. Nevertheless this process is used to show that it is possible to match the target shape with the proposed transformation method. The results are presented in Figure \ref{resultsTrafo} (d). Only the rectangle has a curved edge at the backside.
Since the $\mathcal{L}_{SSE}$ is taking all values of $u$, $v$ and $p$ into account, the prediction near the edges of the objects is getting slightly worse. Thus, the relative error of $\Delta p$ and $F_i$ is increasing compared to the former loss function.
\\
\indent
To validate the performance of the shape optimization, CFD-simulations are done for the transformed geometries with FEniCS. Figure \ref{comparisonPredCFD} presents the velocity component $u$ of two different airfoils for the case where $\Delta p$, $F_i$
and geometric restrictions are used as constraints. The first row displays the predicted values of the U-Net for the transformed geometries. The second row displays the values of the CFD-simulation of the same transformed geometries. It is visible that the predicted values are close to the values from the CFD-simulation. As a comparison the CFD-simulation of the target geometry is revealed in the last row. The values from the transformed geometries are similar to the CFD-simulated values of the target geometry. This is showing that the presented method is able to find a solution which is fulfilling the physical constraints.

\section{Conclusion}
The proposed method demonstrates how objects can be transformed subject to varying physical constraints, thus providing a flexible and transferable model to a wide variety of problem setups. Beside the presented results concerning flow field prediction, this method can be applied to any kind of data representing vector fields, like electromagnetic fields or forces and stress values in a mechanical system. Further this method can also be applied to other tasks that aim to estimate parameters of objects. The next step is to apply the model to complex real applications and extend it to 3D data.






\bibliographystyle{IEEEtran}
\bibliography{IEEEabrv,egbib}

\begin{thebibliography}{10}
\providecommand{\url}[1]{#1}
\csname url@samestyle\endcsname
\providecommand{\newblock}{\relax}
\providecommand{\bibinfo}[2]{#2}
\providecommand{\BIBentrySTDinterwordspacing}{\spaceskip=0pt\relax}
\providecommand{\BIBentryALTinterwordstretchfactor}{4}
\providecommand{\BIBentryALTinterwordspacing}{\spaceskip=\fontdimen2\font plus
\BIBentryALTinterwordstretchfactor\fontdimen3\font minus
  \fontdimen4\font\relax}
\providecommand{\BIBforeignlanguage}[2]{{%
\expandafter\ifx\csname l@#1\endcsname\relax
\typeout{** WARNING: IEEEtran.bst: No hyphenation pattern has been}%
\typeout{** loaded for the language `#1'. Using the pattern for}%
\typeout{** the default language instead.}%
\else
\language=\csname l@#1\endcsname
\fi
#2}}
\providecommand{\BIBdecl}{\relax}
\BIBdecl

\bibitem{planning_trajectories}
S.~Hoppe, Z.~Lou, D.~Hennes, and M.~Toussaint, ``{Planning Approximate
  Exploration Trajectories for Model-Free Reinforcement Learning in
  Contact-Rich Manipulation},'' \emph{IEEE Robotics and Automation Letters},
  vol.~PP, pp. 1--1, 2019.

\bibitem{Raissi2018DeepHP}
M.~Raissi, ``{Deep Hidden Physics Models: Deep Learning of Nonlinear Partial
  Differential Equations},'' \emph{J. Mach. Learn. Res.}, vol.~19, pp.
  25:1--25:24, 2018.

\bibitem{SIRIGNANO20181339}
J.~Sirignano and K.~Spiliopoulos, ``{DGM: A deep learning algorithm for solving
  partial differential equations},'' \emph{Journal of Computational Physics},
  vol. 375, pp. 1339--1364, 2018.

\bibitem{DBLP:journals/corr/abs-1810-08217}
N.~Thuerey, K.~Weissenow, H.~Mehrotra, N.~Mainali, L.~Prantl, and X.~Hu,
  ``{Well, how accurate is it? {A} Study of Deep Learning Methods for
  Reynolds-Averaged Navier-Stokes Simulations},'' \emph{CoRR}, vol.
  abs/1810.08217, 2018.

\bibitem{10.5555/3305890.3306035}
J.~Tompson, K.~Schlachter, P.~Sprechmann, and K.~Perlin, ``{Accelerating
  Eulerian Fluid Simulation with Convolutional Networks},'' in
  \emph{Proceedings of the 34th International Conference on Machine Learning -
  Volume 70}, ser. ICML’17.\hskip 1em plus 0.5em minus 0.4em\relax JMLR.org,
  2017, pp. 3424--3433.

\bibitem{Xie2018tempoGANAT}
Y.~Xie, E.~Franz, M.~Chu, and N.~Th{\"u}rey, ``{tempoGAN: a temporally
  coherent, volumetric GAN for super-resolution fluid flow},'' \emph{ArXiv},
  vol. abs/1801.09710, 2018.

\bibitem{Lagaris_1998}
I.~Lagaris, A.~Likas, and D.~Fotiadis, ``Artificial neural networks for solving
  ordinary and partial differential equations,'' \emph{IEEE Transactions on
  Neural Networks}, vol.~9, no.~5, pp. 987--1000, 1998.

\bibitem{Aarts_2001}
L.~Aarts and P.~van~der Veer, ``{Neural Network Method for Solving Partial
  Differential Equations},'' \emph{Neural Processing Letters}, vol.~14, pp.
  261--271, 2001.

\bibitem{schenck2018spnets}
C.~Schenck and D.~Fox, ``{SPNets: Differentiable Fluid Dynamics for Deep Neural
  Networks},'' \emph{CoRR}, vol. abs/1806.06094, 2018.

\bibitem{Ummenhofer2020Lagrangian}
B.~Ummenhofer, L.~Prantl, N.~Thuerey, and V.~Koltun, ``{Lagrangian Fluid
  Simulation with Continuous Convolutions},'' in \emph{International Conference
  on Learning Representations}, 2020.

\bibitem{10.1145/2939672.2939738}
X.~Guo, W.~Li, and F.~Iorio, ``{Convolutional Neural Networks for Steady Flow
  Approximation},'' in \emph{Proceedings of the 22nd ACM SIGKDD International
  Conference on Knowledge Discovery and Data Mining}, ser. KDD ’16.\hskip 1em
  plus 0.5em minus 0.4em\relax New York, NY, USA: Association for Computing
  Machinery, 2016, pp. 481--490.

\bibitem{wu2019comprehensive}
Z.~Wu, S.~Pan, F.~Chen, G.~Long, C.~Zhang, and P.~S. Yu, ``{A Comprehensive
  Survey on Graph Neural Networks},'' 2019.

\bibitem{DBLP:journals/corr/RonnebergerFB15}
O.~Ronneberger, P.~Fischer, and T.~Brox, ``{U-Net: Convolutional Networks for
  Biomedical Image Segmentation},'' \emph{CoRR}, vol. abs/1505.04597, 2015.

\bibitem{DBLP:journals/corr/JaderbergSZK15}
M.~Jaderberg, K.~Simonyan, A.~Zisserman, and K.~Kavukcuoglu, ``{Spatial
  Transformer Networks},'' \emph{CoRR}, vol. abs/1506.02025, 2015.

\bibitem{NIPS2014_5423}
I.~Goodfellow, J.~Pouget-Abadie, M.~Mirza, B.~Xu, D.~Warde-Farley, S.~Ozair,
  A.~Courville, and Y.~Bengio, ``{Generative Adversarial Nets},'' in
  \emph{Advances in Neural Information Processing Systems 27}, Z.~Ghahramani,
  M.~Welling, C.~Cortes, N.~D. Lawrence, and K.~Q. Weinberger, Eds.\hskip 1em
  plus 0.5em minus 0.4em\relax Curran Associates, Inc., 2014, pp. 2672--2680.

\bibitem{SUN2015415}
G.~Sun, Y.~Sun, and S.~Wang, ``{Artificial neural network based inverse design:
  Airfoils and wings},'' \emph{Aerospace Science and Technology}, vol.~42, pp.
  415--428, 2015.

\bibitem{Nanophotonic_Structures}
D.~Liu, Y.~Tan, and Z.~Yu, ``{Training Deep Neural Networks for the Inverse
  Design of Nanophotonic Structures},'' \emph{ACS Photonics}, vol.~5, 2017.

\bibitem{raissi2018hidden}
M.~Raissi, A.~Yazdani, and G.~E. Karniadakis, ``{Hidden Fluid Mechanics: A
  Navier-Stokes Informed Deep Learning Framework for Assimilating Flow
  Visualization Data},'' 2018.

\bibitem{Kim2018DeepFA}
B.~Kim, V.~C. Azevedo, N.~Thuerey, T.~Kim, M.~H. Gross, and B.~Solenthaler,
  ``{Deep Fluids: A Generative Network for Parameterized Fluid Simulations},''
  \emph{Comput. Graph. Forum}, vol.~38, pp. 59--70, 2018.

\bibitem{Latent_Space_Physics}
S.~Wiewel, M.~Becher, and N.~Thuerey, ``{Latent Space Physics: Towards Learning
  the Temporal Evolution of Fluid Flow},'' \emph{Computer Graphics Forum},
  vol.~38, 2018.

\bibitem{doi:10.2514/1.J057894}
V.~Sekar, M.~Zhang, C.~Shu, and B.~C. Khoo, ``{Inverse Design of Airfoil Using
  a Deep Convolutional Neural Network},'' \emph{AIAA Journal}, vol.~57, no.~3,
  pp. 993--1003, 2019.

\bibitem{AlnaesBlechta2015a}
M.~S. Aln{\ae}s, J.~Blechta, J.~Hake, A.~Johansson, B.~Kehlet, A.~Logg,
  C.~Richardson, J.~Ring, M.~E. Rognes, and G.~N. Wells, ``{The FEniCS Project
  Version 1.5},'' \emph{Archive of Numerical Software}, vol.~3, no. 100, 2015.

\bibitem{kingma2014adam}
D.~P. Kingma and J.~Ba, ``{Adam: A Method for Stochastic Optimization},'' 2014.

\bibitem{24792}
F.~L. Bookstein, ``{Principal warps: thin-plate splines and the decomposition
  of deformations},'' \emph{IEEE Transactions on Pattern Analysis and Machine
  Intelligence}, vol.~11, no.~6, pp. 567--585, 1989.

\bibitem{10.1007/3-540-47977-5_2}
G.~Donato and S.~Belongie, ``{Approximate Thin Plate Spline Mappings},'' in
  \emph{Computer Vision --- ECCV 2002}, A.~Heyden, G.~Sparr, M.~Nielsen, and
  P.~Johansen, Eds.\hskip 1em plus 0.5em minus 0.4em\relax Berlin, Heidelberg:
  Springer Berlin Heidelberg, 2002, pp. 21--31.

\end{thebibliography}
%




\end{document}